%% file: icra_2024.tex
\documentclass[letterpaper, 10 pt, conference]{ieeeconf}  

\IEEEoverridecommandlockouts                              

\usepackage{graphicx}
\usepackage{epsfig} 
\usepackage{mathptmx} 
\usepackage{times} 
\usepackage{amsmath} 
\usepackage{amssymb}  

\usepackage{subcaption}
\usepackage{multirow}
\usepackage{hyperref}
\usepackage{xcolor}
\usepackage{cite}
\usepackage{colortbl}
\usepackage{dirtytalk}
\usepackage[normalem]{ulem}

\newcommand{\rulesep}{\unskip\ \vrule\ }

\title{\LARGE \bf
A Framework For Automated Dissection Along Tissue Boundary
}

\author{Ki-Hwan Oh$^{1}$,  Leonardo Borgioli$^{1}$,  Milo\v s \v Zefran$^{1}$, Liaohai Chen$^{2}$ and Pier Cristoforo Giulianotti$^{2}$
\thanks{*This work was not supported by any organization}
\thanks{$^{1}$K.H. Oh, L. Borgioli and Milo\v s \v Zefran are with the Robotics Lab,  Department of Electrical and Computer Engineering, College of Engineering, University of Illinois Chicago, Chicago, IL 60607, USA.}%
\thanks{$^{2}$L. Chen and P.C. Giulianotti are with the Surgical Innovation and Training Lab,  Department of Surgery, College of Medicine, University of Illinois Chicago, Chicago, IL 60607, USA.}%
}

\begin{document}

\maketitle
\thispagestyle{empty}
\pagestyle{empty}

\begin{abstract}

Robotic surgery promises enhanced precision and adaptability over traditional surgical methods. It also offers the possibility of automating surgical interventions, resulting in reduced stress on the surgeon, better surgical outcomes, and lower costs. Cholecystectomy, the removal of the gallbladder, serves as an ideal model procedure for automation due to its distinct and well-contrasted anatomical features between the gallbladder and liver, along with standardized surgical maneuvers. Dissection is a frequently used subtask in cholecystectomy where the surgeon delivers the energy on the hook to detach the gallbladder from the liver. Hence, dissection along tissue boundaries is a good candidate for surgical automation. For the da Vinci surgical robot to perform the same procedure as a surgeon automatically, it needs to have the ability to (1) recognize and distinguish between the two different tissues (e.g. the liver and the gallbladder), (2) understand where the boundary between the two tissues is located in the 3D workspace, (3) locate the instrument tip relative to the boundary in the 3D space using visual feedback, and (4) move the instrument along the boundary. This paper presents a novel framework that addresses these challenges through AI-assisted image processing and vision-based robot control. We also present the ex-vivo evaluation of the automated procedure on chicken and pork liver specimens that demonstrates the effectiveness of the proposed framework.


\end{abstract}


\input{introduction}

\input{methodology}

\input{results}

\input{conclusion}

\newpage
\nocite{*}
\bibliographystyle{IEEEtran}
\bibliography{icra_2024}
\newpage
\input{appendix}

\end{document}

%% file: introduction.tex
\section{Introduction}

Robotic-assisted surgery (RAS) has emerged as a transformative approach in recent years, offering distinct advantages over traditional open and laparoscopic procedures. Among the myriad of surgical operations, cholecystectomy, the surgical removal of the gallbladder, stands out as an ideal candidate for automation. Its relatively straightforward surgical anatomy, combined with standardized maneuvers and well-defined anatomical features, makes it well-suited for studies aiming to automate surgical procedures using robots. 

Recent work in RAS automation includes ablation~\cite{DBLP:journals/corr/AyvaliSWRSC15}, blunt dissection~\cite{blunt_dissection}, and several works on suturing~\cite{Jackson2016,Lu2022autosuture,iyer2013single,sen2016suture,Tracking_thread}. For instance, \cite{Sagitov2018} presents a detection and planning algorithm for automated suturing using visual tracking of suture threads. Challenges and solutions for robot-assisted knot tying are investigated in~\cite{9664632}, emphasizing the role of image guidance in enhancing the precision of suturing tasks.

The cameras are the main sensing modality in RAS, and considerable research has been done on processing these images. In~\cite{doi:10.1126/scitranslmed.aad9398}, creating point clouds allows for a 3D modeling of the surgical site. These models can then be used to guide the robot's movements.
The ability to generate and process point clouds are especially crucial when dealing with deformable objects, as it provides a dynamic representation that can adapt to changes in real-time. However, the point clouds in~\cite{doi:10.1126/scitranslmed.aad9398} are created from plenoptic cameras, which are not present in most operating rooms. Furthermore, the addition of a second camera to the surgical setup is not compatible with the current robot-assisted laparoscopy, which relies on endoscopic images. Other studies have employed visual segmentation models to detect and define loose connective tissues, aiding surgeons in visualizing safe dissection planes during procedures like robot-assisted gastrectomy~\cite{Kumazu2021}. Nevertheless, the processing is at the pixel level, and there is no segmentation and logical connection to individual tissues.

Compared to auto-suturing, there seems to be less work on auto-dissection, indicating a potential research gap~\cite{doi:10.1146/annurev-control-062420-090543}. However, \cite{autocutting} introduces a vision-based cutting control algorithm, emphasizing the role of visual servoing in enhancing the precision of automated dissection. This algorithm allows for the automatic cutting of deformable objects along a predetermined path with a scalpel, utilizing visual feedback from feature points for real-time parameter estimation. In RAS, surgeons never use scalpels due to bleeding and use energy delivery with monopolar, bipolar, or ultrasonic instruments instead.

\input{figures/sys_arch}

In our work, we use the da Vinci surgical robot with the da Vinci Research Kit (dVRK)~\cite{dvrk}. Our research focuses on vision-based robot control tailored to tissue dissection\footnote{In the surgical literature, dissection refers to the actual separation of tissues; in our work, the term refers to energy delivery to a particular location on a specimen.}. Our main contributions are: (a) We introduce a novel comprehensive framework for dissecting the tissue relying solely on endoscopic images, that is, what is provided to a surgeon to control the robot. (b) To support precise robot control, we developed an ArUco-based~\cite{GARRIDOJURADO20142280} calibration for dVRK, which is easy to implement and remains feasible when the configuration of the da Vinci system is changed. (c) We augmented the dVRK to control the power supply that delivers energy to the monopolar and bipolar instruments for dissection. This allows us to use additional instruments like the Permanent Cautery Hook (PCH), which is rarely used in research but widely used in cholecystectomy. (d) We generated a custom dataset that categorizes different types of tissues and da Vinci instruments. This dataset was used for fine-tuning Detectron2~\cite{wu2019detectron2}, and we show that the new model can be used in RAS for segmenting tissues and detecting keypoints of instruments. This allows the system to identify the boundary between different tissues and locate the instrument. (e) We conducted an ex-vivo evaluation of our proposed system using pig liver and chicken specimens.

We do not claim that we have achieved realistic tissue dissection as we do not address tissue flexibility or use a grasper to pull the tissue during the dissection; these are left for future work. However, we show that we can successfully automate a number of fundamental subtasks during tissue dissection and, for the first time, demonstrate a robotic system that can autonomously deliver energy along tissue boundaries. Our work is, therefore, an important step towards autonomous dissections that approach surgeon's skill.

%% file: figures/sys_arch.tex


\begin{figure*}[t]
    \centering
    \includegraphics[scale=0.57]{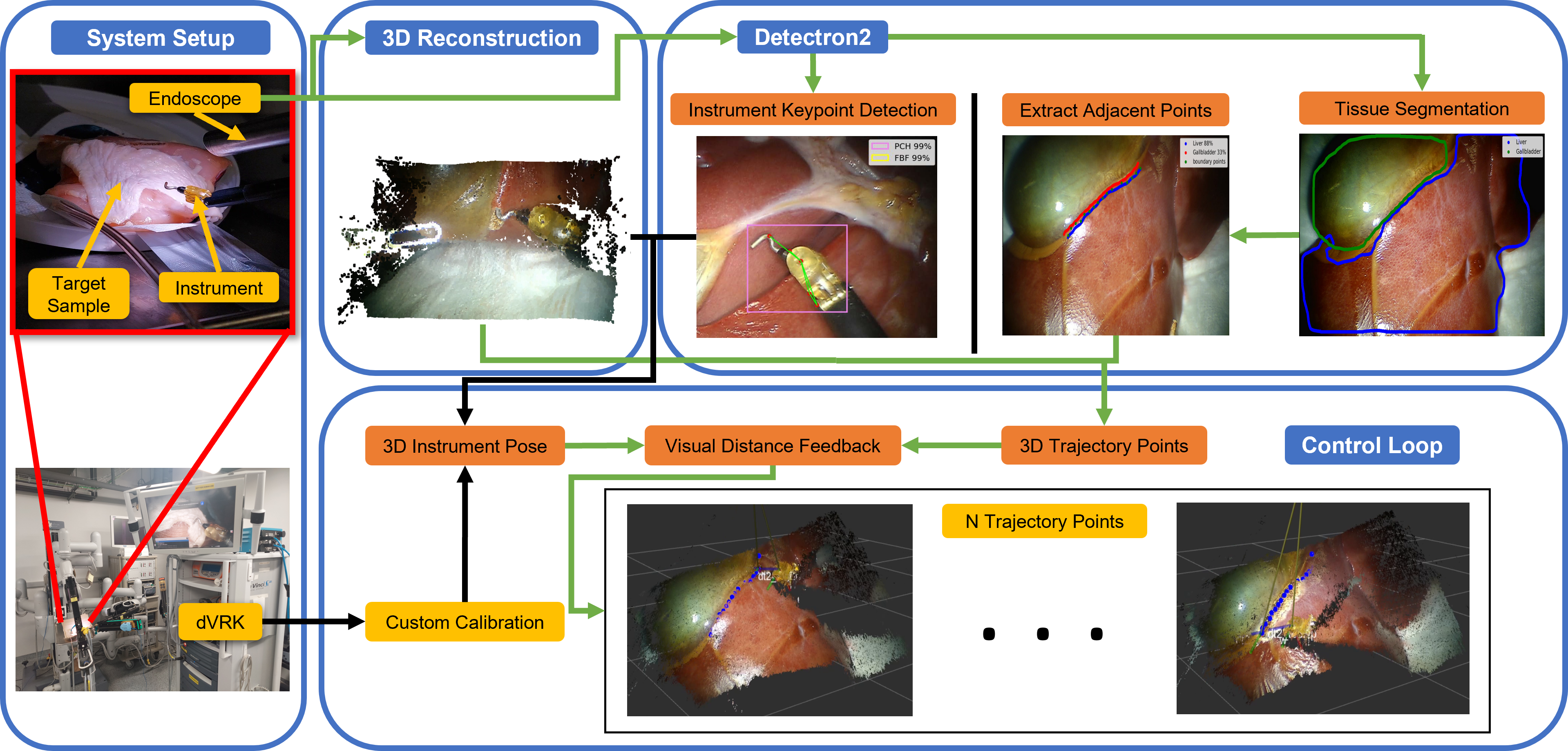}
    \caption{The overall system architecture: (a) Hardware setup. (b) Results from Detectron2, both segmentation and key point detection. (c) The extracted goal points and tooltip pose in the 3D space along with the inputs used by the control system.}
    \label{fig:sys_arch}
\vspace{-6mm}
\end{figure*}

%% file: methodology.tex
\section{Methodology}

\subsection{Hardware Setup}

This work uses the first-generation da Vinci system with the integration of dVRK. In a departure from conventional configurations, we opted for the Si model endoscope, which is distinguished by its superior image quality and notably reduced noise characteristics. The dVRK lacks direct control over the external power supply that governs the voltage output for the monopolar and bipolar instruments used for dissecting the tissues. To address this, we interfaced the energy device cables and the console pedals with an Arduino, enabling us to manage the signals for both monopolar and bipolar instrument operations.

\subsection{Custom Calibration}

In the full da Vinci system, the forward kinematics of the dVRK are derived from the Setup Joints (SUJs), the base of the da Vinci robot.
This allows the Patient Side Manipulator (PSM) tip's configuration to be computed with respect to the Endoscope Camera Manipulator (ECM) tip. Notably, the positional discrepancy between the PSM tip and the ECM tip remains within a range of $+/- 5cm$ for translation and between $5\sim10$ degrees for orientation~\cite{dvrk}. While such an error margin may be acceptable for manual robot operation using Master Tool Manipulators (MTMs), it becomes critical during automated control. As a solution, we introduced a custom calibration for the dVRK utilizing an external camera and fiducial markers.

Our approach is influenced by~\cite{camera_calib_dvrk}, which calibrates the dVRK using an optical tracking system
equipped with custom adapters for instrument tips. Given the limited availability of such a system, we opted for the more accessible ArUco markers~\cite{GARRIDOJURADO20142280} and a ZED mini external camera. Additionally, we selected base frames for each arm that offer greater flexibility, ensuring consistent performance during real surgical procedures, as depicted in Fig.~\ref{fig:calibsetup}. Note that the base frames remain visible even when a tray is present in the da Vinci workspace. In our work, we use the Product of Exponentials~\cite{murray1994robot} formalism for forward kinematics. We thus have: 
\begin{equation}
g_{rt}(\vartheta) = e^{\hat{\zeta}_{1}\vartheta_{1}}e^{\hat{\zeta}_{2}\vartheta_{2}}e^{\hat{\zeta}_{3}\vartheta_{3}}e^{\hat{\zeta}_{4}\vartheta_{4}}e^{\hat{\zeta}_{5}\vartheta_{5}}e^{\hat{\zeta}_{6}\vartheta_{6}}g_{rt}(0)
\end{equation}
\begin{equation}
g_{se}(\varphi) = e^{\hat{\xi}_{1}\varphi_{1}}e^{\hat{\xi}_{2}\varphi_{2}}e^{\hat{\xi}_{3}\varphi_{3}}e^{\hat{\xi}_{4}\varphi_{4}}g_{se}(0)
\end{equation}
where $g_{ab}$ represents the transformation (homogeneous matrix) between frames $A$ and $B$. Frames $R$ and $S$ are the base frames for the PSM and ECM, respectively, while $T$ and $E$ represent their corresponding instrument tip frames (Fig.~\ref{fig:calibsetup}). Vectors $\zeta$ and $\xi$ (in $\mathbb{R}^6$) are the unknown joint twist parameters, and $\vartheta$ and $\varphi$ are the joint angles for PSM and ECM derived from dVRK measurements, respectively.

\input{figures/calib_setup}

We can assume that the original dVRK console joint angles $\theta_{d}$ are computed as:
\begin{equation} 
\theta_{d} = \alpha_{d}\epsilon_{d} + \beta_{d}
\end{equation}
where $\epsilon_{d}$ is the raw encoder reading while $\alpha$ and $\beta$ are constant (linear) parameters. We would like to compute the (calibrated) joint angle $\theta_{c}$ from the raw encoder reading according to:
\begin{equation}
\begin{split}
\theta_{c} &= \alpha_{c}\epsilon_{d} + \beta_{c}= \alpha^{\prime}\theta_{d} + \beta^{\prime} 
\end{split}
\label{Eq:Angles}
\end{equation}
where $\alpha^{\prime}=\alpha_{c}/\alpha_{d}$ and $\beta^{\prime} = \beta_{c}-\alpha^{\prime}\beta_{d}$. Hence, we linearly map the dVRK's joint angles to calibrated joint angles. 

We collected the dataset for calibration by sweeping each joint from its minimum to maximum position while holding the rest of the joints at zero, and the position of the instrument ArUco markers was averaged over 10 frames for accuracy at each configuration, momentarily stopping the robot. We then utilized MATLAB's Global Optimization Toolbox~\cite{matlabgo} to determine the joint twist parameters as well as the parameters in~\eqref{Eq:Angles} from this dataset. Further details are described in~\ref{sec:armcalib}.

\input{figures/customcalib}

\input{tables/table_calibration}

To evaluate our custom forward kinematics, we recorded the custom kinematics when moving the arms to random positions as in Fig.~\ref{fig:customcalib}. After determining the $g_{rt}$ and $g_{se}$, we can establish a relative configuration of the PSM tip wrt. the ECM tip, incorporating the Helper ($H$) frame as seen in Fig.~\ref{fig:calibsetup}.The Helper frame may seem unnecessary when using only a single arm. However, it becomes essential when adding a second PSM. This is because both the ECM and the first PSM's base frame are occluded from the other side. The Helper frame facilitates visibility and connection between the second PSM and the ECM. If the SUJs' locations are altered, only the transformation between the helper and base frames needs updating.

\begin{equation}
g_{et} = g_{se}^{-1}\cdot g_{hs}^{-1}\cdot g_{hr}\cdot g_{rt} = g_{es}\cdot g_{sh}\cdot g_{hr}\cdot g_{rt}
\end{equation}

\subsection{Image Segmentation and Keypoint Detection}

The core challenge in automating surgical procedures is accurately extracting surgical instrument locations and locations of target tissues from the endoscopic images. 
To achieve this, we first curated a custom dataset employing the Segment Anything Model (SAM), which was pre-trained with the Segment Anything 1 Billion (SA-1B)~\cite{kirillov2023segany}. 
It utilizes prompts to define the segmentation targets in an image, which is suitable for a broad spectrum of segmentation tasks without requiring further training. Manually selected points were provided as prompts, which are categorized as either positive or negative to indicate whether they belong to the foreground (the object of interest) or the background.
The project couldn't use existing segmentation datasets such as CholecSeg8k ~\cite{hong2020cholecseg8k}, or CholecTriplet2021 ~\cite{nwoye2023cholectriplet2021} because they're based on in-vivo human liver surgeries, while this study involves ex-vivo porcine models. Differences in anatomy and surgical environment between humans and pigs, as well as in-vivo and ex-vivo settings, affect tissue and instrument appearance. Furthermore, providing a streamlined workflow that easily generalizes to other tissues and procedures is an important contribution to our work.

Keypoint annotations were performed manually using COCO annotator~\cite{cocoannotator}. The selection of keypoints for each instrument captures discriminative features and ensures consistency, as key points should remain invariant to common transformations. The selected points are: 
\begin{itemize}
    \item \textbf{TipRight/TipLeft} located for the Fenestrated Bipolar Forceps (FBF) and Large Needle Driver (LND) at the two extremities of the grippers and for the Permanent Cautery Hook (PCH) at the tip and beginning of the hook.
    \item \textbf{TipCenter} located on the center screw of all the instruments.
    \item \textbf{\textbf{Edge}} located on the fifth joint for the FBF and LND or the left side screw of the PCH
    \item \textbf{Head} located on the top screw of all the instruments. 
\end{itemize}

\input{figures/seg_sample}

\input{figures/kpt_sample}

\input{tables/customdata}

These keypoints are on key parts of the tools with different colors and distinctive edges to maximize their detection. 
We recorded endoscopic images while performing motions with both arms. To ensure data diversity, the endoscope angle and the position of the ex-vivo material (chicken or pig liver) were changed several times, resulting in a total of 2820 images. Details of the number of instances annotated for segmentation and keypoint detection are described in Table~\ref{tab:customdata}. The dataset adheres to Microsoft’s COCO format~\cite{mscoco}, ensuring compatibility and ease of integration.

As shown in the first two rows of Fig.~\ref{fig:seg_sample}, from semi-automatically annotated images for image segmentation and manual annotations for keypoint detection (Fig.~\ref{fig:kpt_anns}) it was then possible to obtain a precise (cf. Table~\ref{tab:dt2res}) model generalizable to different shapes of specimens (pig livers and chicken) and instrument configurations.  

The annotated dataset was trained on Detectron2~\cite{wu2019detectron2}, a state-of-the-art object segmentation and keypoint detection model. The training was based on the pre-trained mask R-CNN model R50-FPN since it had the fastest inference time among different models with similar performance. The learning rate was initialized as 0.05 with a warm-up phase of 1000 iterations. The model was trained for 9000 iterations to ensure comprehensive learning, and the learning rate was reduced by $20\%$ every 300 iterations. Additionally, a batch size of 512 for the Region of Interests aids in capturing more object instances. These hyperparameters represent a balanced approach to achieving accurate and efficient image segmentation, taking into account computational constraints and dataset characteristics. Similar hyperparameters have been used for keypoint detection.

\subsection{3D Scene Reconstruction}

A significant component of our methodology is the 3D reconstruction of the surgical scene. Before the reconstruction, we captured multiple images, including a $9\times6$ chessboard in different positions and orientations. The cameras were calibrated with selected low reprojection error images according to the traditional approach in~\cite{zhang2000flexible} from the MATLAB Stereo Camera Calibration Toolbox and OpenCV~\cite{opencv_library}. Subsequently, using the intrinsic and extrinsic camera parameters, we applied the modified Semi-global Matching algorithm (SGM)~\cite{sgm} to produce stereo disparity maps from the stereo endoscopic images. The images are passed through a bilateral filter before applying SGM to reduce the noise but maintain the edges as much as possible.
The disparity map is then converted to point clouds using the baseline and the focal length of the stereo cameras.
We average each 3D tissue boundary and instrument position point within a fixed-size ($10\times10$) window, thereby identifying points representing infinite distances and eliminating them from the disparity map as outliers.

\subsection{Control Algorithm}

Our approach leverages Detectron2 for processing endoscopic images, differentiating between dual targets like muscle and skin in chicken samples. The adjacent points were found by the traditional method for finding nearest neighbors with K-Dimensional Tree (KD Tree)~\cite{songrit1999kdtree}. Subsequently, the edge point between each set of background tissue (e.g., liver) and the primary tissue (e.g., gallbladder) trajectory point was extracted based on the peak disparity value between the two points. Afterward, the edge points were downsampled based on the farthest-first traversal algorithm~\cite{rosenkrantz1977analysis}.

Furthermore, Detectron2 identifies instrument keypoints within endoscope frames. The corresponding 3D positions of the collected 2D image points are determined from the generated point clouds. The robot is initially programmed to move $0.5mm$ toward the first trajectory point in each iteration loop. Based on the keypoint detection results, the robot tracks the distance between the current trajectory point and the instrument tip's 3D position. Once the deviation falls below a $0.5mm$ threshold, indicating the target's been reached, the system shifts its goal to the next trajectory point. This control mechanism is termed Position-Based Visual Servoing (PBVS), where the feedback relies on the 3D information from the stereo images.

%% file: figures/calib_setup.tex
\begin{figure}[t]
    \centering
    \includegraphics[width=\linewidth]{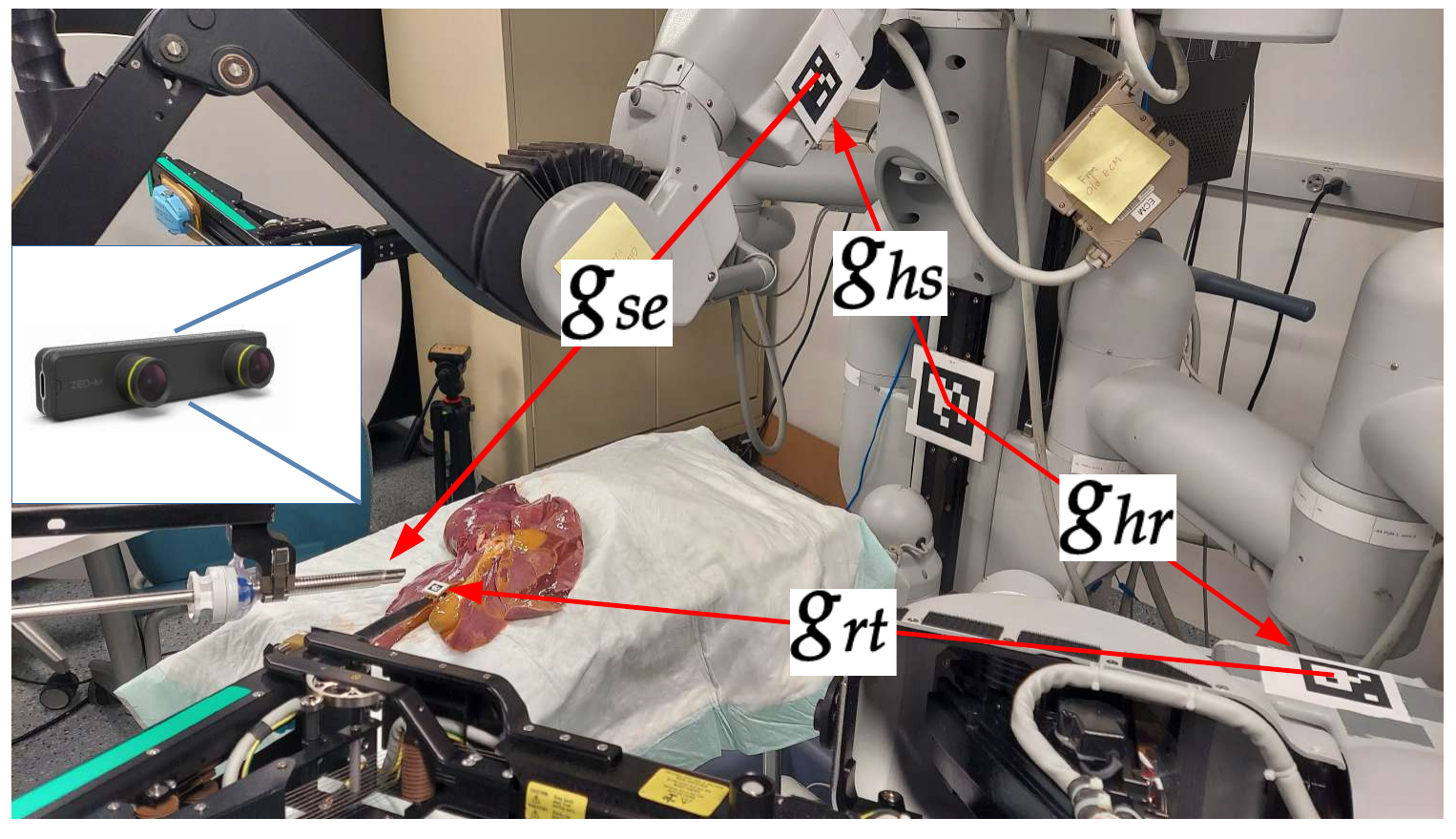}
    \caption{The setup showing how our custom-calibrated kinematics work. The transformations are shown based on the direction of the arrows and eventually, they are used to find the transformation between the ECM tip and PSM tip.}
    \label{fig:calibsetup}
\vspace{-3mm}
\end{figure}

%% file: figures/customcalib.tex
\begin{figure}[t]
    \centering
    \includegraphics[width=0.95\linewidth]{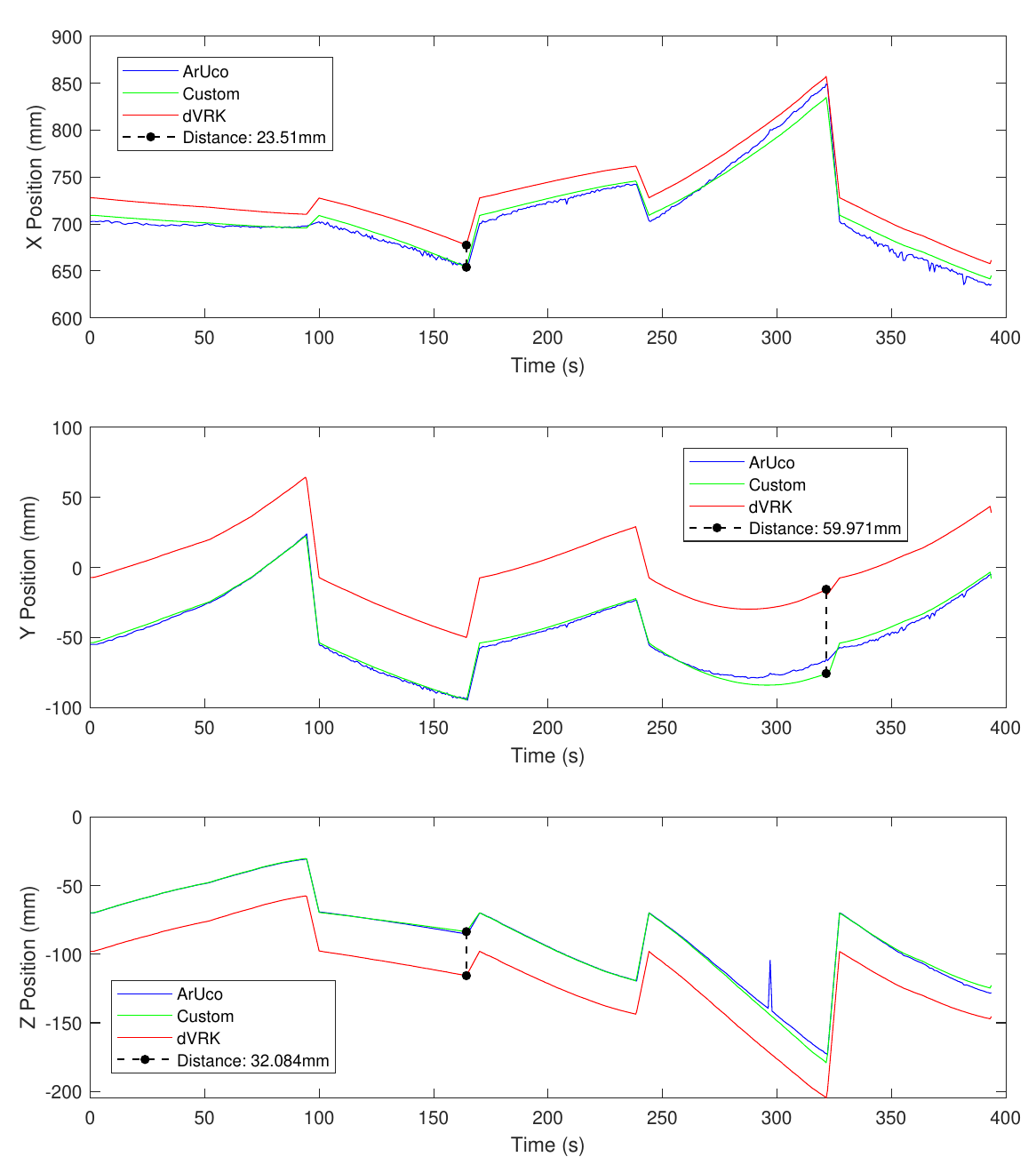}
    \caption{Position predicted by the dVRK forward kinematics compared to the ArUco marker, and the calibrated kinematics when tested during a random motion.}
    \label{fig:customcalib}
\vspace{-6mm}
\end{figure}

%% file: tables/table_calibration.tex
\begin{table}[t]
\centering
\begin{tabular}{l|l|l}
\hline
\hline
\textbf{Measure (distance)}          & \textbf{dVRK $\Leftrightarrow$ ArUco }     & \textbf{Custom $\Leftrightarrow$ ArUco} \\ \hline

\multirow{3}{*}{Mean Error}          &X: $11.0mm$                                &X: $5.6mm$                               \\
                                     &Y: $50.1mm$                                &Y: $2.5mm$                               \\
                                     &Z: $20.9mm$                                &Z: $1.1mm$                               \\ \hline
\multirow{3}{*}{Standard deviation}  &X: $4.9mm$                                &X: $3.6mm$                               \\
                                     &Y: $2.0mm$                                &Y: $2.1mm$                               \\
                                     &Z: $3.9mm$                                &Z: $2.2mm$                               \\
\hline
\hline
\end{tabular}
\caption{The first row shows the distance between the kinematics data obtained by dVRK compared to the one obtained by detecting the ArUco markers, while the second row shows the distance between the kinematics data by our custom calibration compared to the one obtained by detecting the ArUco markers. We can observe that the mean error decreases drastically (a factor of 1.96 for X, 20.04 for Y, and 19 for Z). }
\label{tab:calib_table}
\vspace{-6mm}
\end{table}

%% file: figures/seg_sample.tex
\begin{figure}[t]
    \begin{subfigure}[t]{0.49\columnwidth}
        \centering
        \includegraphics[width=\linewidth]{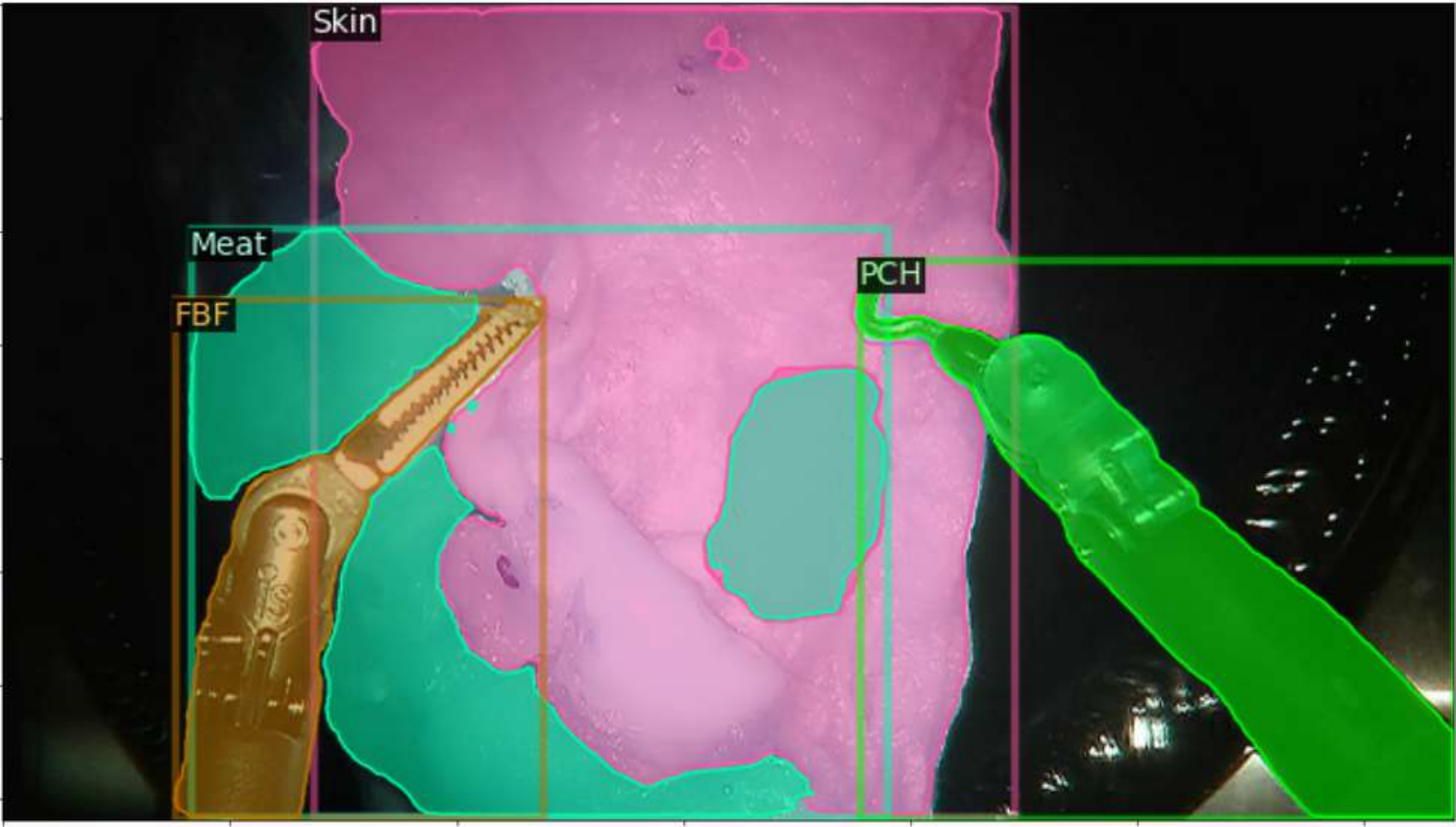}
        \includegraphics[width=\linewidth]{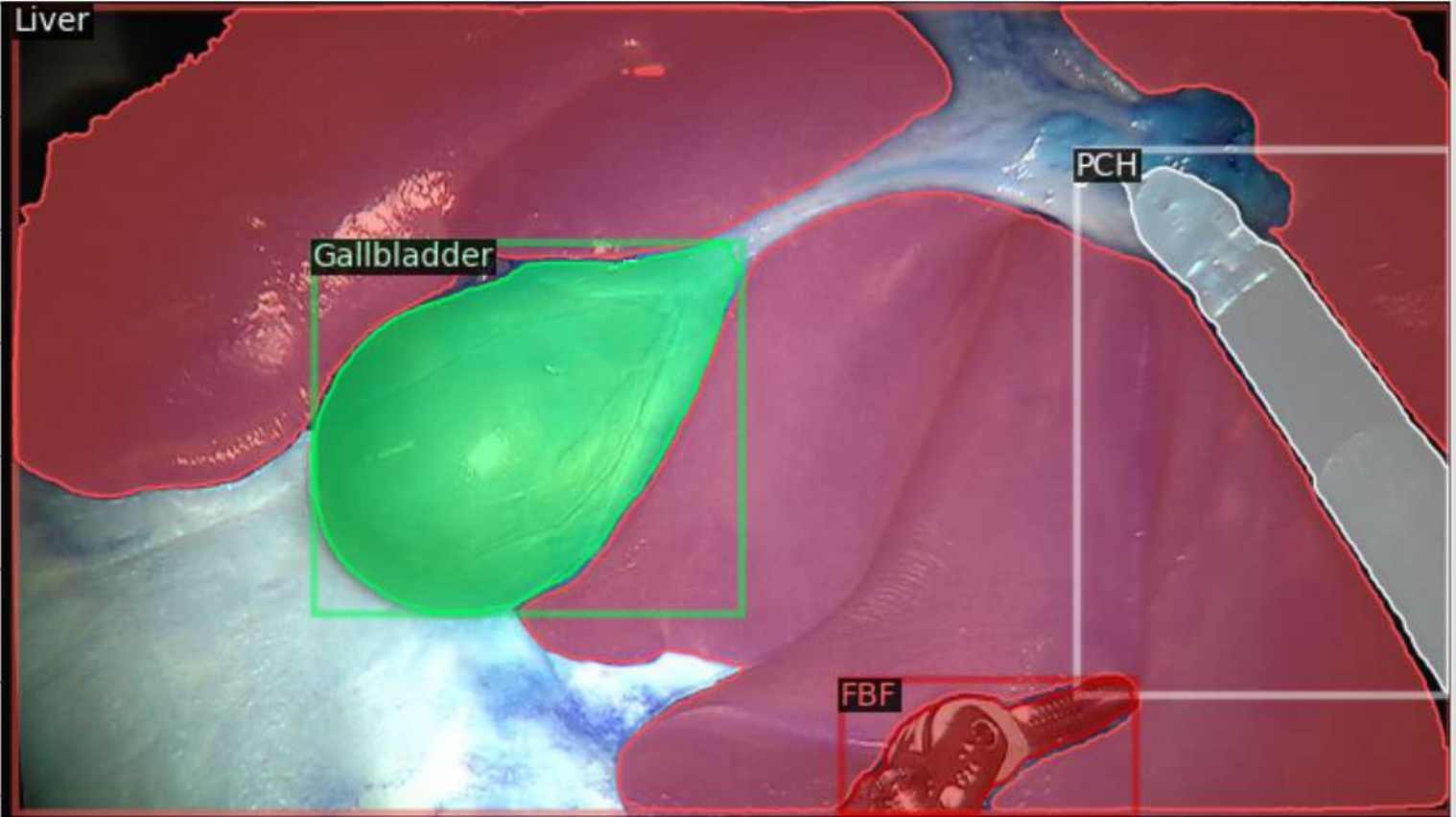}
        \caption{Annotations}
        \label{fig:annotations}
    \end{subfigure}
    \hfill
    \begin{subfigure}[t]{0.49\columnwidth}
        \centering
        \includegraphics[width=\linewidth]{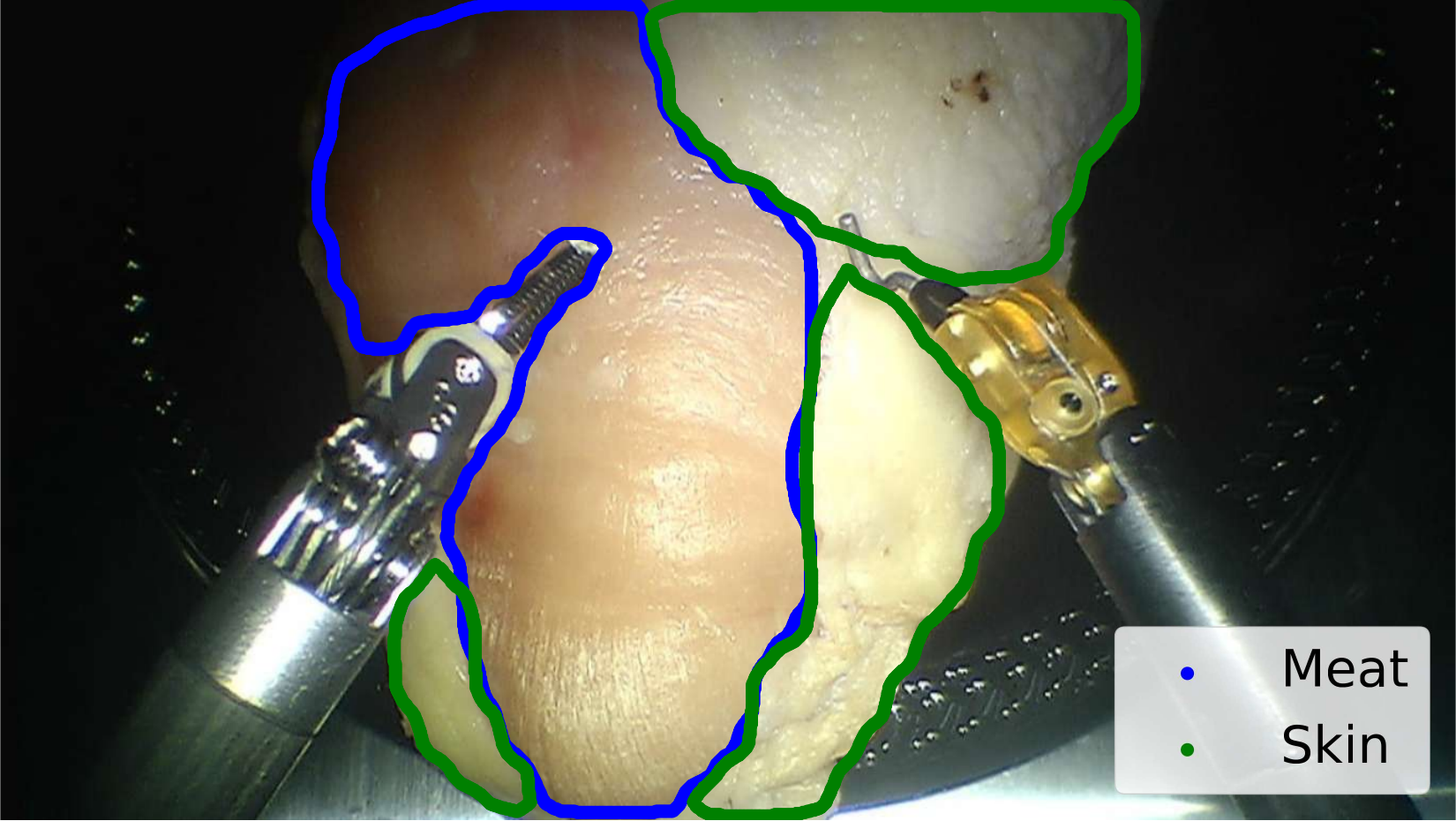}
        \includegraphics[width=\linewidth]{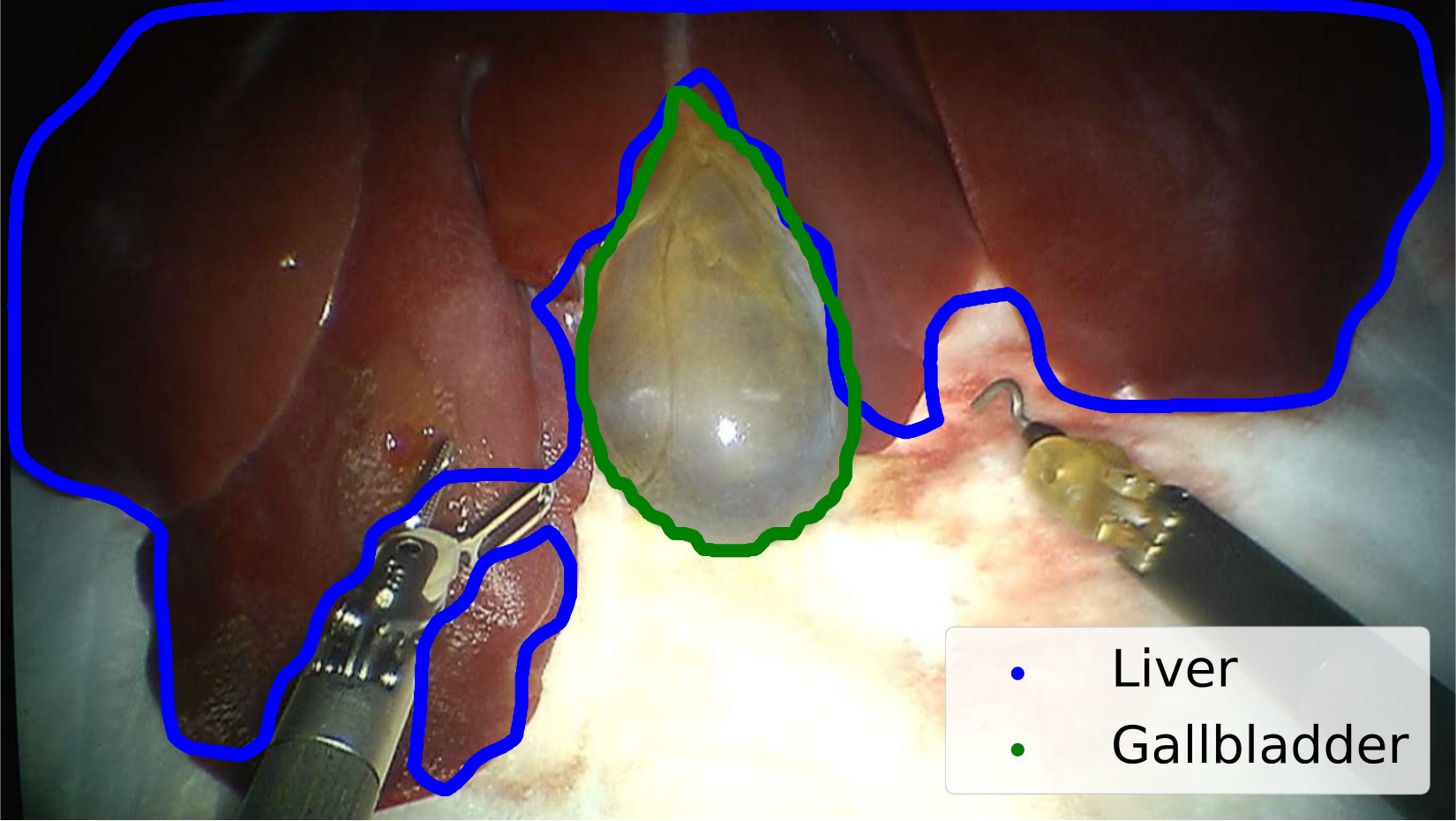}
        \caption{Predictions}
        \label{fig:predictions}
    \end{subfigure}
\caption{(a) Samples of the manually annotated segmentation dataset with SAM. (b) Samples of segmentation predictions with the trained Detectron2 model.}
\label{fig:seg_sample}
\vspace{-3mm}
\end{figure}

%% file: figures/kpt_sample.tex
\begin{figure}[t]
    \begin{subfigure}[t]{0.49\columnwidth}
        \centering
        \includegraphics[width=\linewidth]{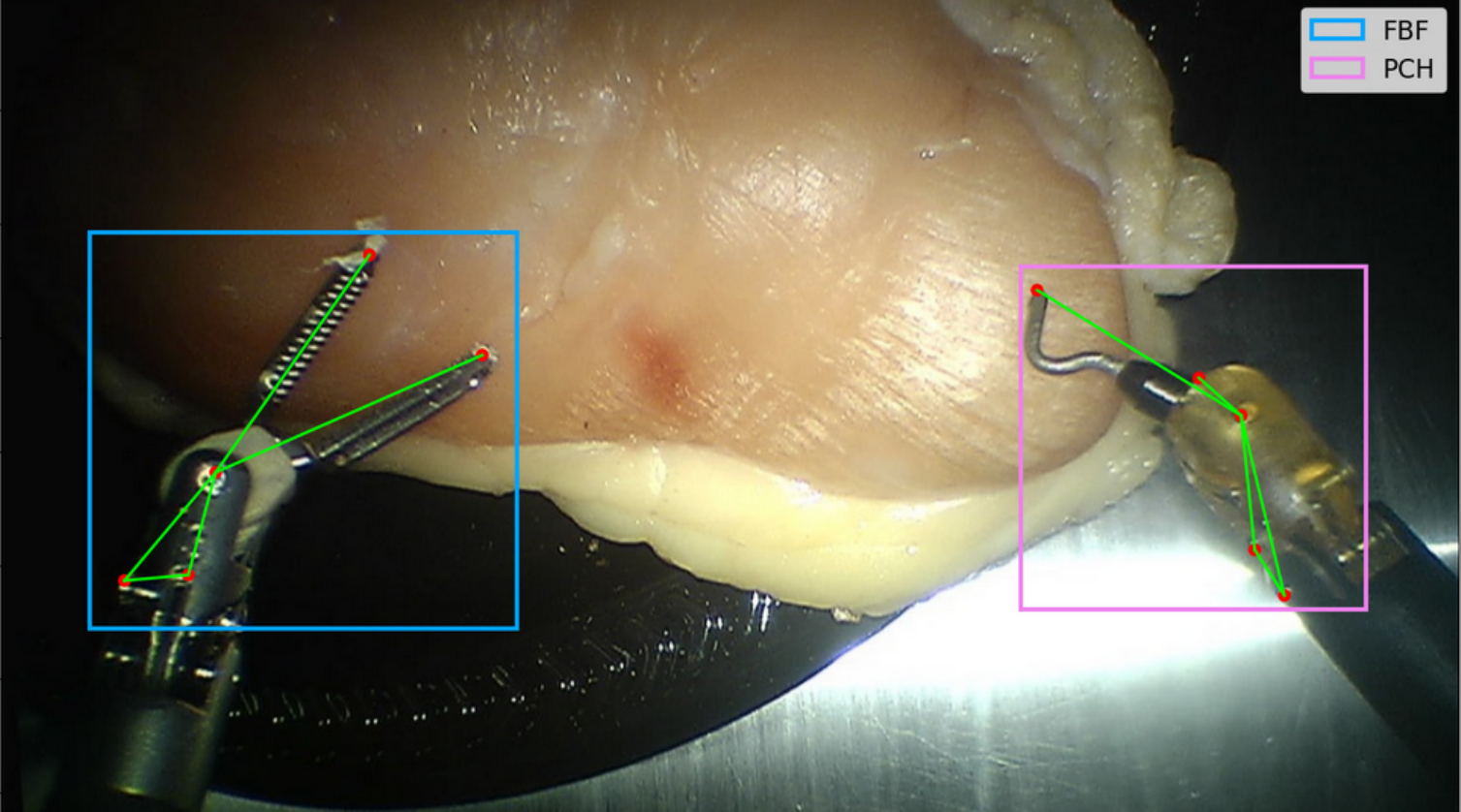}
        \caption{Annotations}
        \label{fig:kpt_anns}
    \end{subfigure}
    \hfill
    \begin{subfigure}[t]{0.49\columnwidth}
        \centering
        \includegraphics[width=\linewidth]{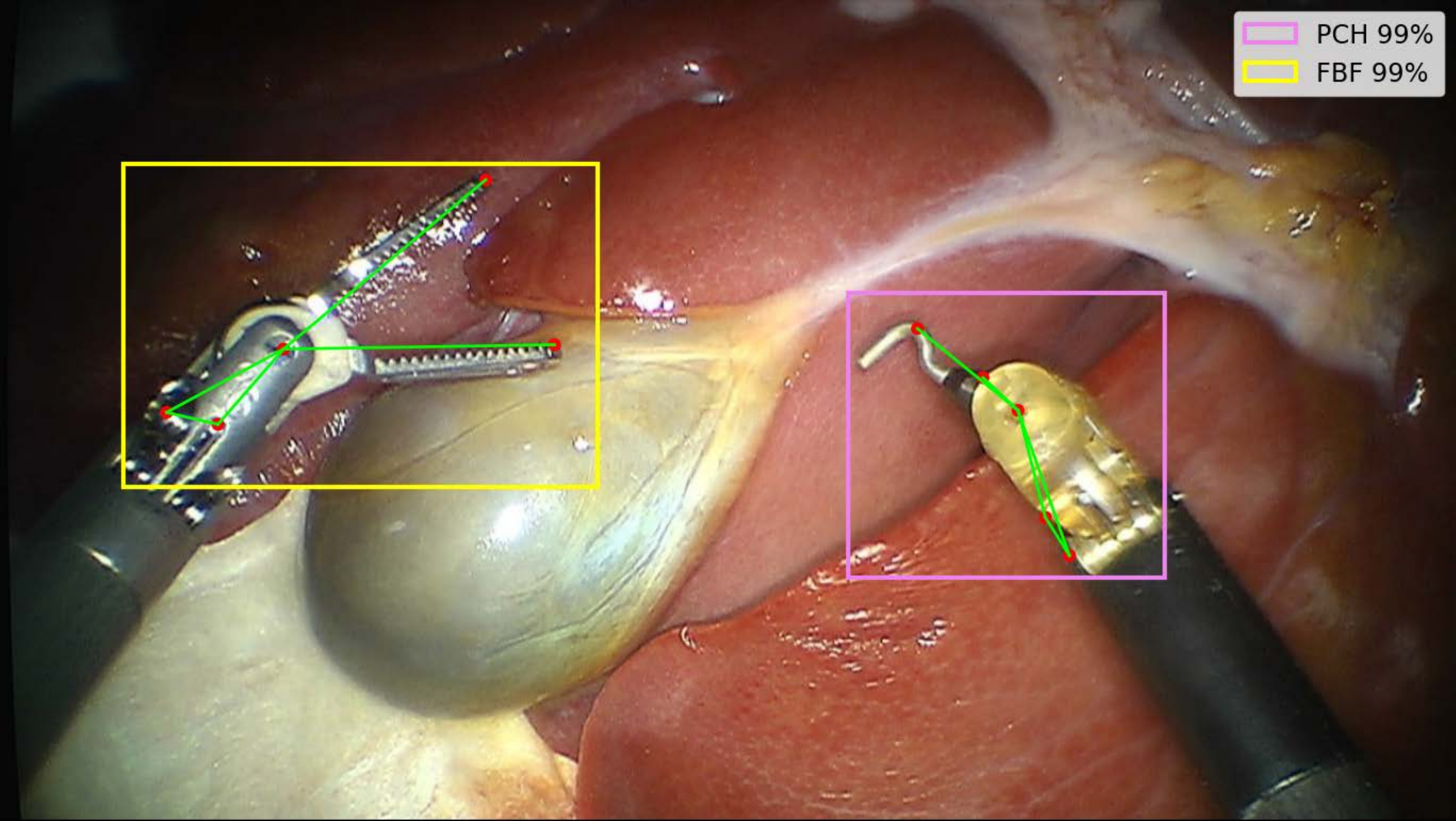}
        \caption{Predictions}
        \label{fig:kpt_preds}
    \end{subfigure}
\caption{(a) Example of the manually annotated keypoint dataset. (b) Example of keypoint predictions by the trained Detectron2 model.}
\label{fig:kpt_sample}
\vspace{-6mm}
\end{figure}

%% file: tables/customdata.tex
\begin{table}[t]
\centering
\begin{tabular}{l|l|c|c}
\hline
\hline
\textbf{Data Type}                     & \textbf{Categories}      & \textbf{Train instances} & \textbf{Test instances} \\ \hline
\multirow{4}{*}{Segmentation} & Chicken Meat    & 1390                    & 348                    \\
                              & Chicken Skin    & 1364                    & 343                    \\
                              & Pig Liver       & 1430                    & 356                    \\
                              & Pig Gallbladder & 1429                    & 359                    \\ \hline
\multirow{3}{*}{Keypoints}    & LND             & 1188                    & 277                    \\
                              & FBF             & 471                     & 130                    \\
                              & PCH             & 1904                    & 477                    \\
\hline
\hline
\end{tabular}
\caption{The number of annotations for each train and test dataset with SAM. Keypoints were annotated for the following instruments: Large Needle Driver (LND), Fenestrated Bipolar Forceps (FBF), and Permanent Cautery Hook (PCH).}
\label{tab:customdata}
\vspace{-6mm}
\end{table}

%% file: results.tex
\section{Results and Discussion}

\subsection{Detectron2 Model Performance Assesment}

\input{tables/dt2res}

We utilized the Average Precision (AP) metric to gauge its efficacy, a widely accepted standard for evaluating object detection models. Our findings reveal that while the model adeptly localized tissues, there were occasional challenges in achieving complete tissue segmentation. Notably, the model precisely differentiated and segmented various tissues within the area of interest. 

The bounding box accuracy for instruments' keypoints had some variability, attributed to inconsistencies in the annotated bounding boxes. These annotations primarily served to provide a contextual reference for keypoint locations. Nevertheless, the model was remarkably accurate in detecting keypoints. The AP scores for each category further confirmed the model's robust performance, with exceptionally high scores observed for keypoints associated with instruments like the LND, FBF, and PCH.


\subsection{Ex-vivo Evaluation}

\input{tables/trajectory_mean_std}

\input{figures/auto_diss_res}

Our initial experiments utilized chicken specimens, leveraging the clear tissue boundaries between skin and muscle as a proxy for the distinction between gallbladder and liver. Following our chicken trials, we replicated the procedure with pig livers that still had gallbladders attached. This varied specimen allowed us to thoroughly evaluate our framework across different tissue contexts and anatomical configurations.

The dissection process was successfully executed for each chicken and liver sample (Fig.~\ref{fig:auto_diss_res} and Fig.~\ref{fig:3dtrajs}). We measured the distance between each recorded position of the instrument tip with the expected ideal trajectory of the instrument, traveling between trajectory points linearly. The results are displayed in Table~\ref{tab:exvivores}, showing submillimeter precision on average.

\input{figures/3dTracks/3dtracks}

While the accuracy of the dissection was very good, we observed two major limitations. First, issues with tissue segmentation arose when encountering tissues with colors (such as yellow or white for the gallbladder) that differed from its typical color (dark green). This imperfection comes from the limited scope of the preliminary dataset, which was sampled from single video footage for each tissue. In addition, the keypoint detection encountered difficulties identifying the correct features when the instrument was at the periphery of the endoscopic image. This was evident when transitioning from point 5 to point 6 in Fig.~\ref{fig:chicken_traj1} and from point 3 to point 5 in Fig.~\ref{fig:liver_traj2}. Such challenges arise from the inherent characteristics of endoscopic cameras, where illumination is concentrated at the center, diminishing towards the edges. Additionally, the dataset used for training only had a few images with instruments on the periphery. We also note that since we did not allow endoscope motion and we only used a single arm, leading to a limited workable volume, we only performed a partial dissection of tissue boundaries rather than of the entire anatomical structure.

%% file: tables/dt2res.tex
\begin{table}[t]
\centering
\begin{tabular}{l|c|c|c}
\hline\hline
\textbf{Categories}      & \textbf{AP (Box)} & \textbf{AP (Seg.)} & \textbf{AP (Keypoints)} \\ \hline
Chicken Meat    & 91.3            & 78.6            & -             \\
Chicken Skin    & 86.1            & 61.3            & -             \\
Pig Liver       & 97.4            & 37.5            & -             \\
Pig Gallbladder & 89.2            & 94.3            & -             \\ \hline
LND             & 81.0            & -                 & 99.0        \\
FBF             & 77.1            & -                 & 94.6        \\
PCH             & 74.2            & -                 & 98.4        \\
\hline\hline
\end{tabular}
\caption{The Average Precision (AP) scores (percentages) for each category (Box stands for Bounding Box and Seg. for Segmentation).}
\label{tab:dt2res}
\vspace{-6mm}
\end{table}

%% file: tables/trajectory_mean_std.tex
\begin{table}
\centering
\begin{tabular}{cc|c|c|c}
\hline \hline
\multicolumn{2}{c|}{\textbf{Trial}} & \textbf{Mean (mm)}    & \textbf{Std. dev. (mm)} & \textbf{Duration (s)} \\ \hline
\multicolumn{1}{c|}{\multirow{6}{*}{Liver}}   & 1 & 0.31 & 0.23 & 128\\
\multicolumn{1}{c|}{}                         & 2 & 0.37 & 0.27 &138\\
\multicolumn{1}{c|}{}                         & 3 & 0.29 & 0.24 &102\\
\multicolumn{1}{c|}{}                         & 4 & 0.35 & 0.28 &113\\
\multicolumn{1}{c|}{}                         & 5 & 0.30 & 0.20 &117\\
\multicolumn{1}{c|}{}                         & 6 & 0.33 & 0.19 &135\\ \hline
\multicolumn{1}{c|}{\multirow{5}{*}{Chicken}} & 1 & 0.28 & 0.18 & 75\\
\multicolumn{1}{c|}{}                         & 2 & 0.59 & 0.72 & 108\\
\multicolumn{1}{c|}{}                         & 3 & 0.35 & 0.25 & 98\\
\multicolumn{1}{c|}{}                         & 4 & 0.41 & 0.35 &153\\
\multicolumn{1}{c|}{}                         & 5 & 0.27 & 0.17 &130\\ \hline
\multicolumn{2}{c|}{Weighted}                     & 0.36 & 0.34 & 121.6\\
\hline \hline
\end{tabular}

\caption{Mean and standard deviation of distances between the recorded hook position and the optimal trajectory path ( i.e. a linear movement between each trajectory point).}

\label{tab:exvivores}
\vspace{-3mm}
\end{table}


%% file: figures/auto_diss_res.tex
\begin{figure}[t]
    \centering
    \begin{subfigure}[t]{0.48\columnwidth}
        \centering
        \includegraphics[width=\linewidth]{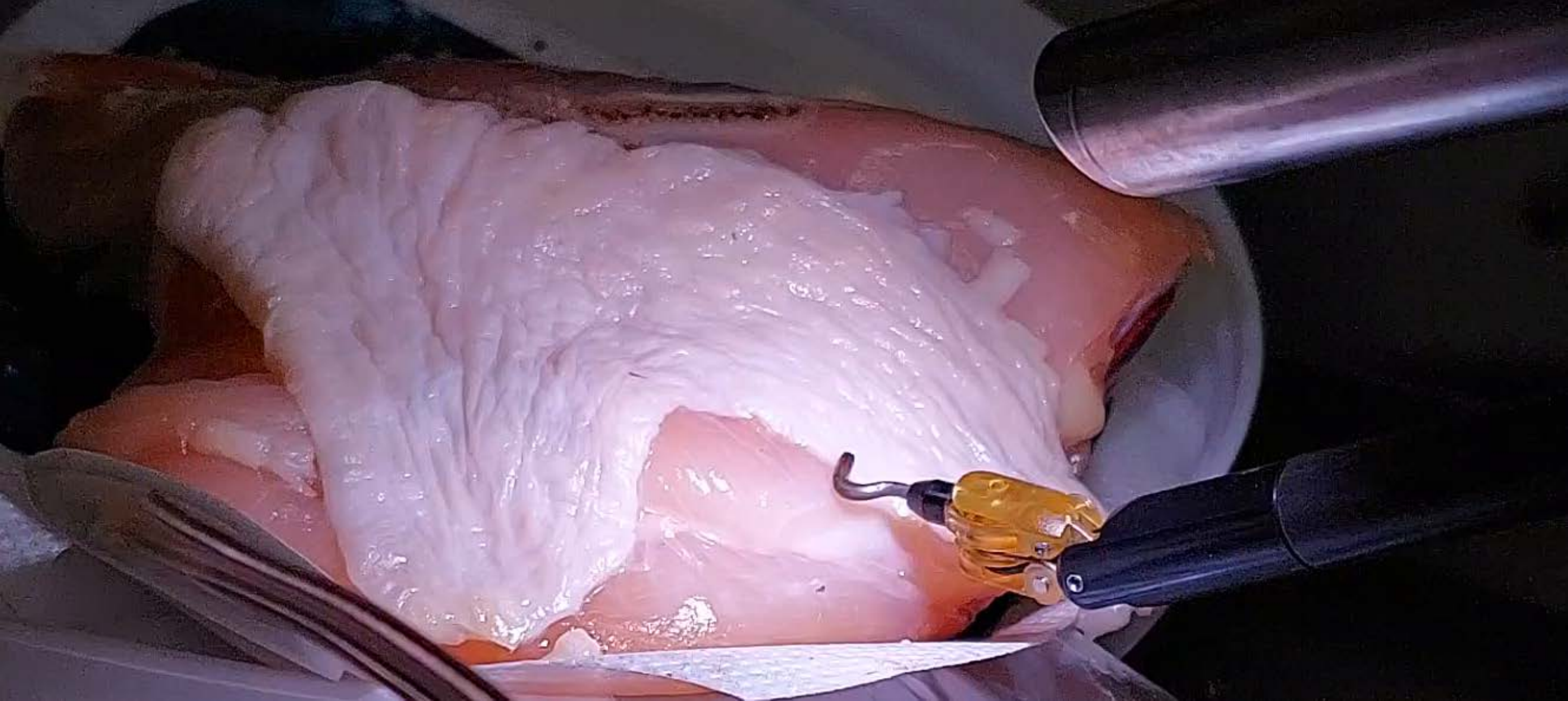}
        \includegraphics[width=\linewidth]{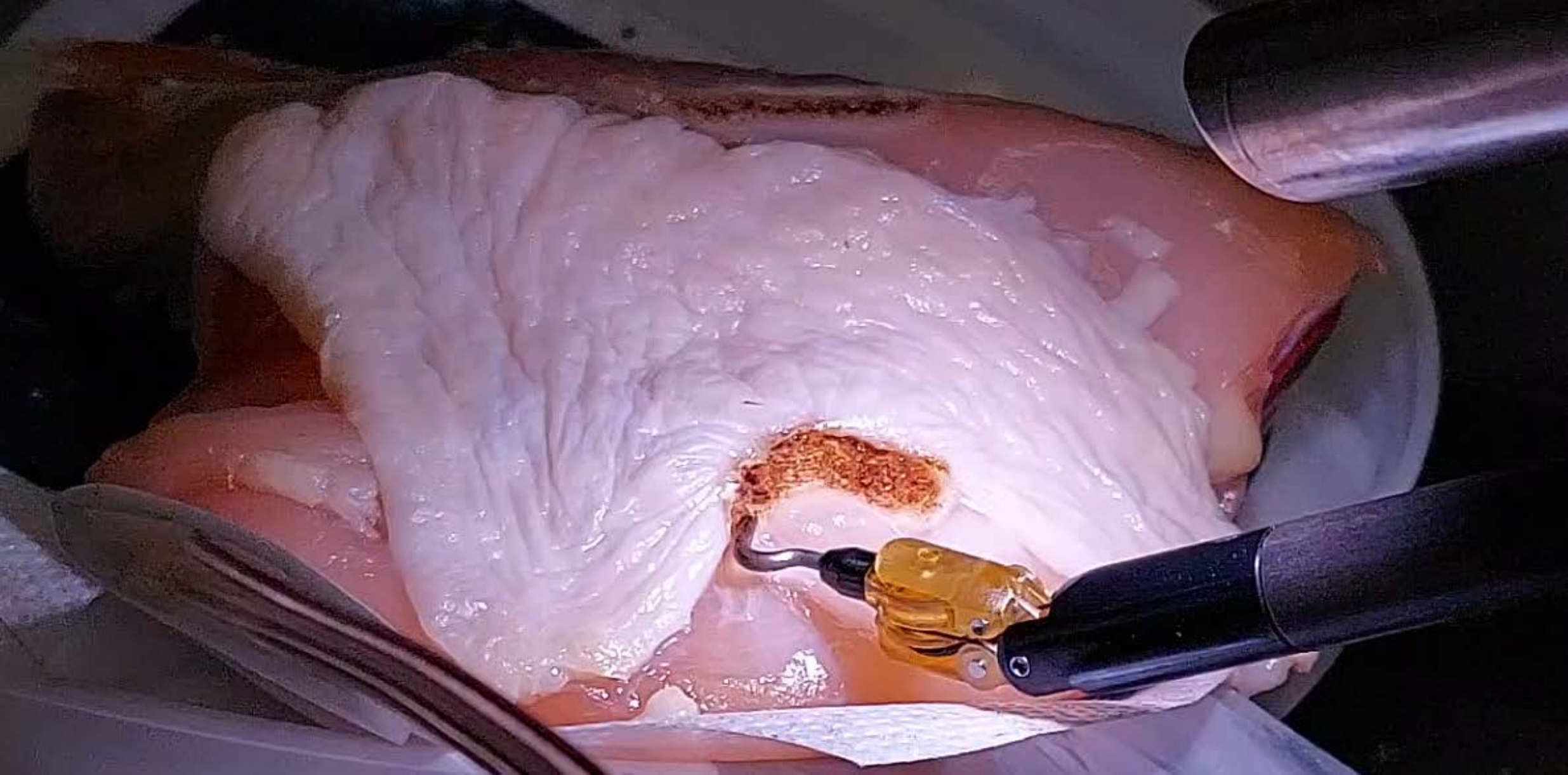}
        \caption{}
        \label{fig:chicken_ann}
    \end{subfigure}
    \begin{subfigure}[t]{0.48\columnwidth}
        \centering
        \includegraphics[width=0.9\linewidth]{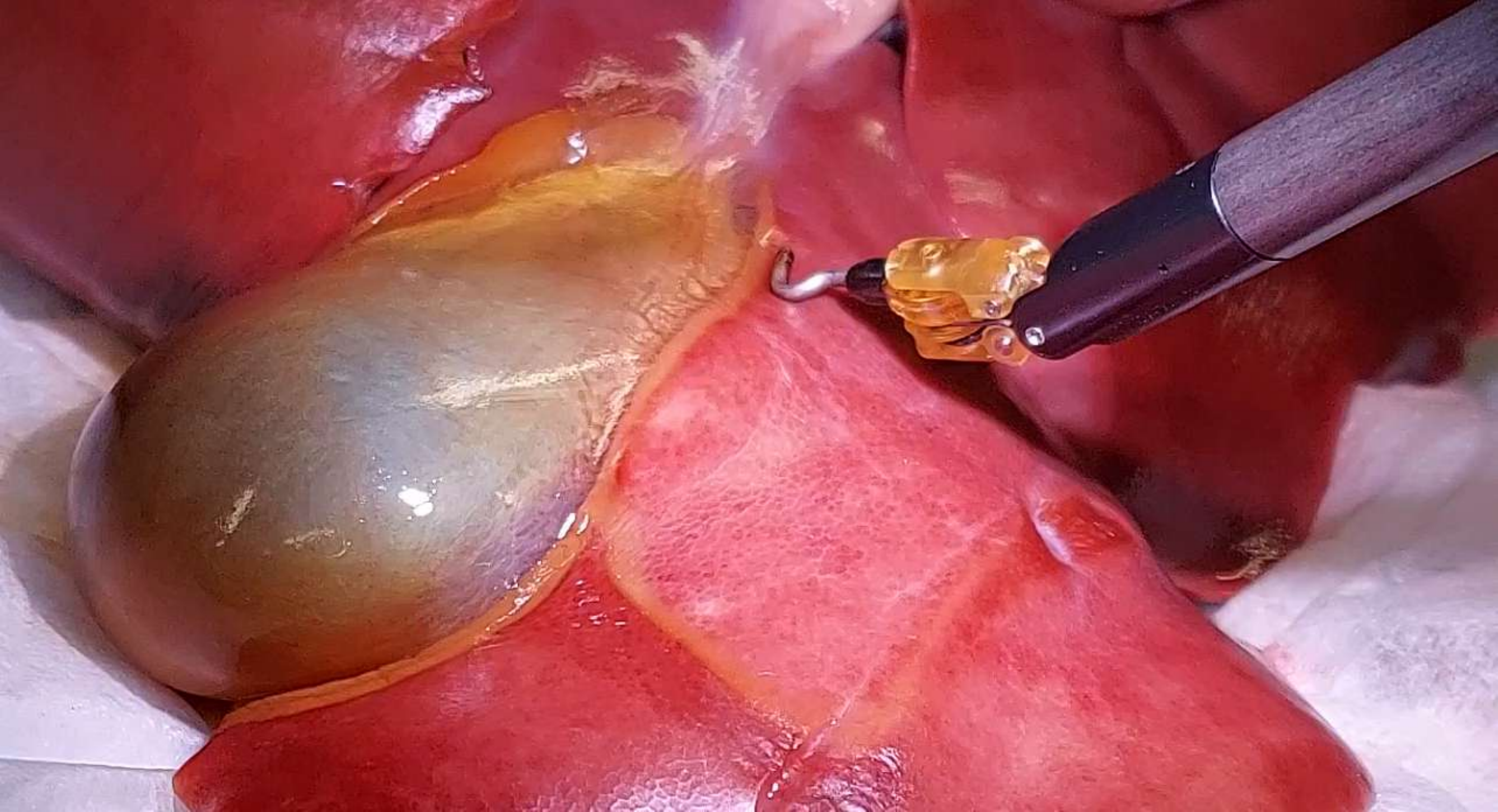}
        \includegraphics[width=0.9\linewidth]{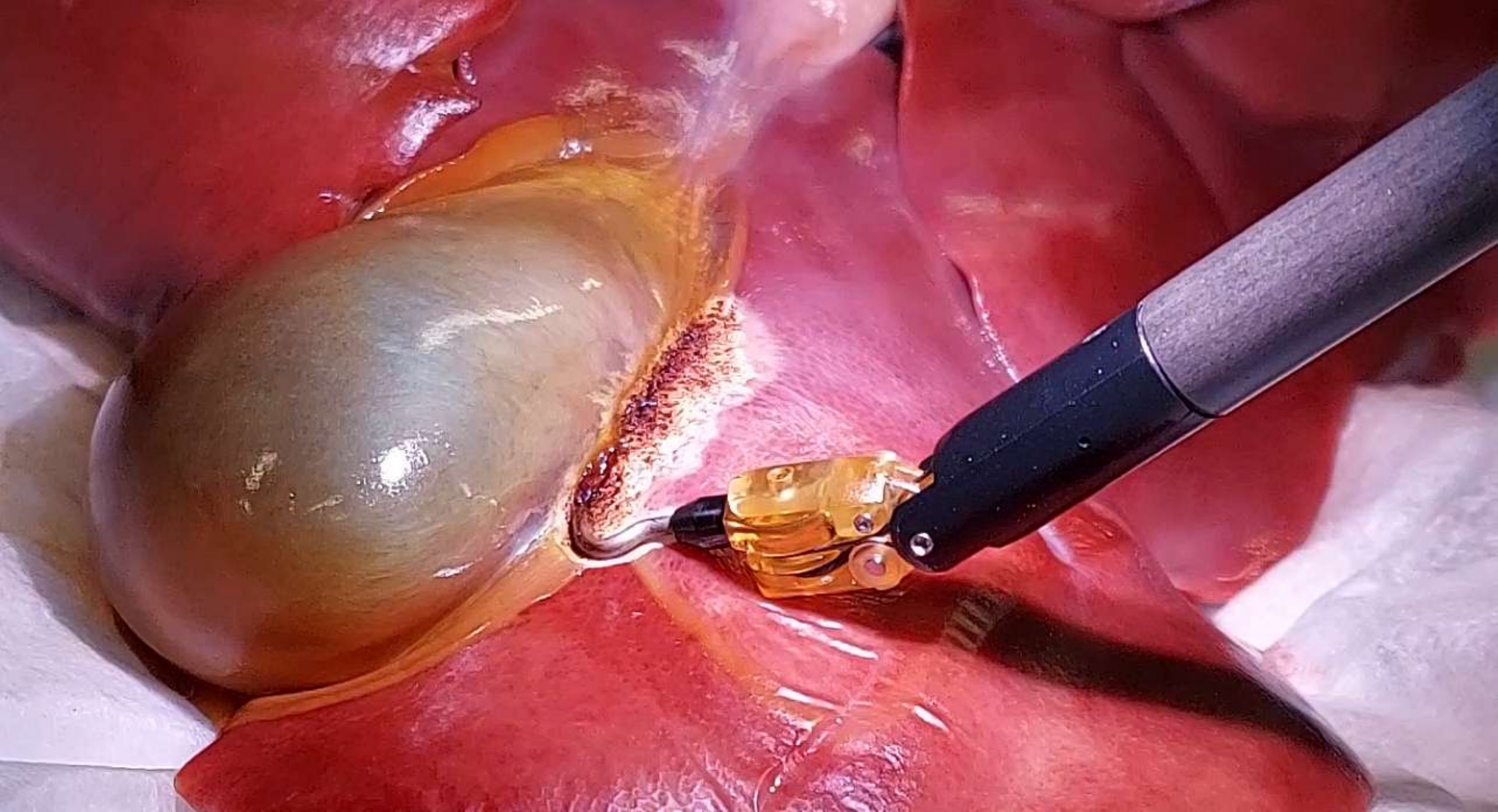}
        \caption{}
        \label{fig:liver_ann}
    \end{subfigure}
    \hfill
    \caption{The images on top show when the instrument reached the first goal point and the ones below show when the instrument reached the final goal point. (a) Energy delivery (monopolar hook) on Chicken, (b) Energy delivery (monopolar hook) on Pig Liver.}
    \label{fig:auto_diss_res}
\vspace{-6mm}
\end{figure}


%% file: figures/3dTracks/3dtracks.tex
\begin{figure*}
    \centering
    \begin{subfigure}[t]{0.3\textwidth}
        \centering
        \includegraphics[width=\linewidth]{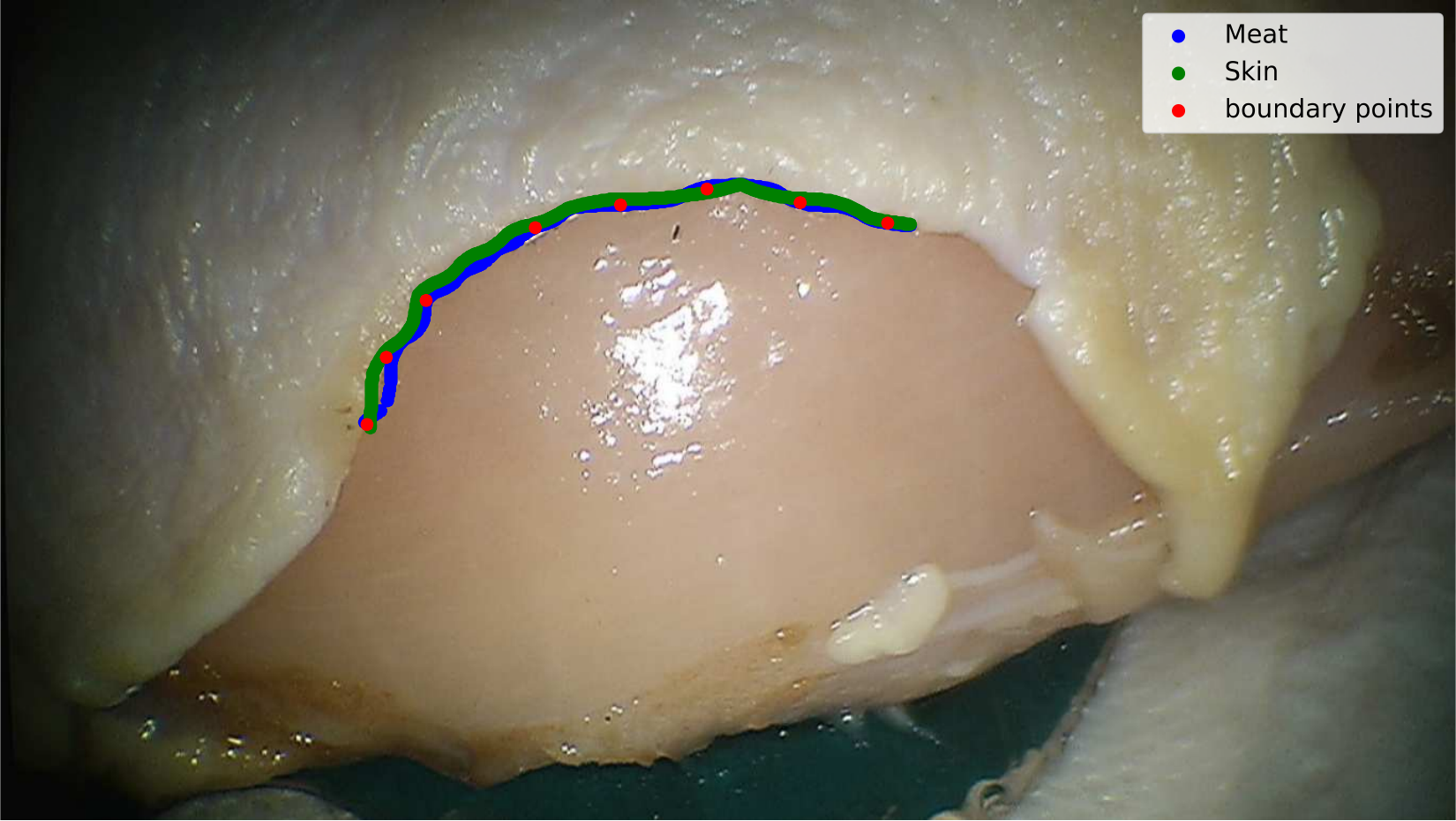}
        \caption{}
        \label{fig:chicken_bnd_pts1}
    \end{subfigure}
    \rulesep
    \hfill
    \begin{subfigure}[t]{0.3\textwidth}
        \centering
        \includegraphics[width=\linewidth]{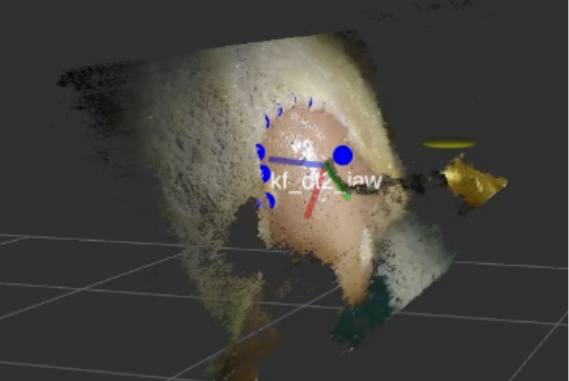}
        \caption{}
        \label{fig:chicken_pcl1}
    \end{subfigure}
    \rulesep
    \hfill
    \begin{subfigure}[t]{0.3\textwidth}
        \centering
        \includegraphics[height=0.2\textheight]{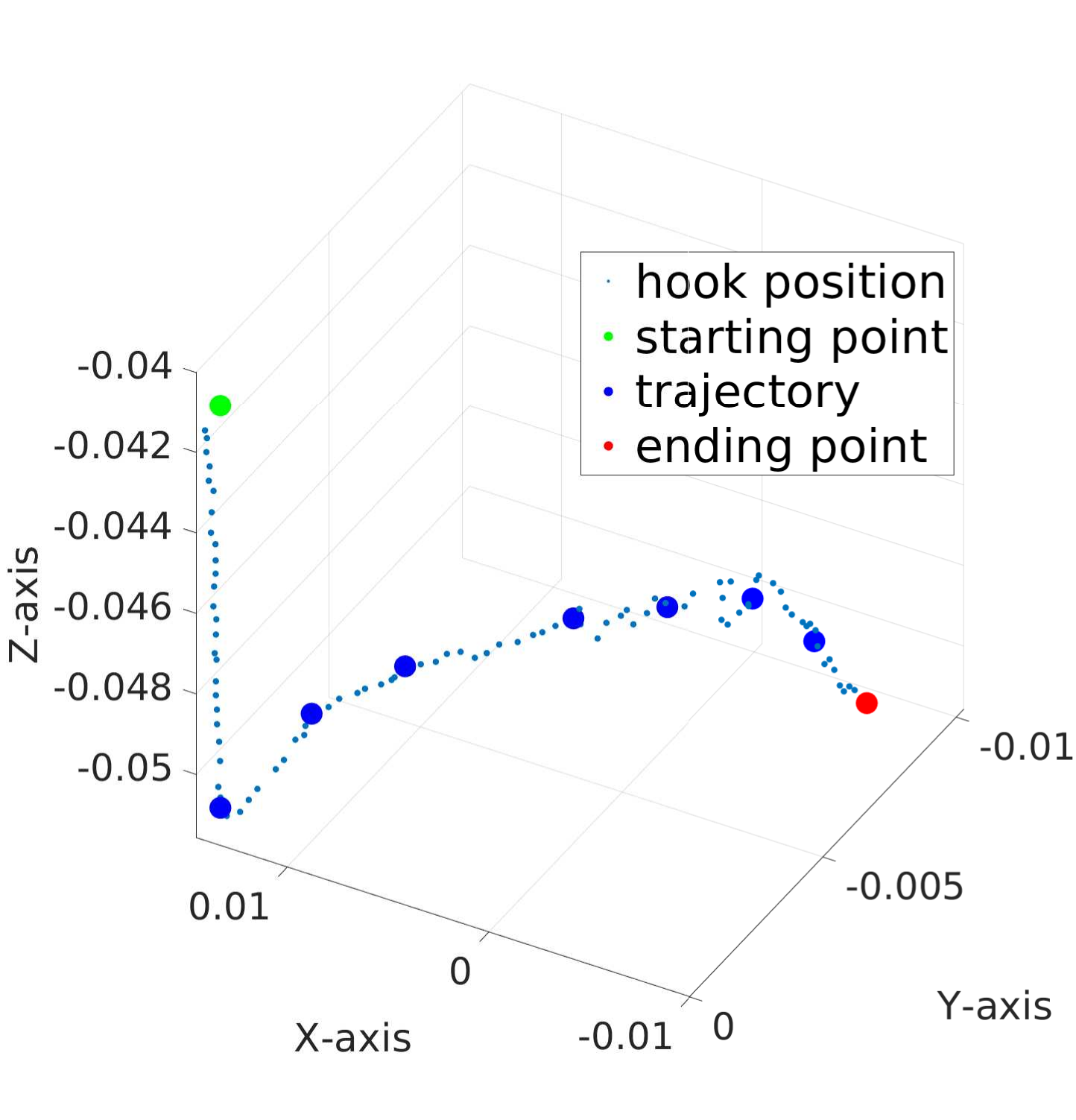}
        \caption{}
        \label{fig:chicken_traj1}
    \end{subfigure}

    \begin{subfigure}[t]{0.3\textwidth}
        \centering
        \includegraphics[width=\linewidth]{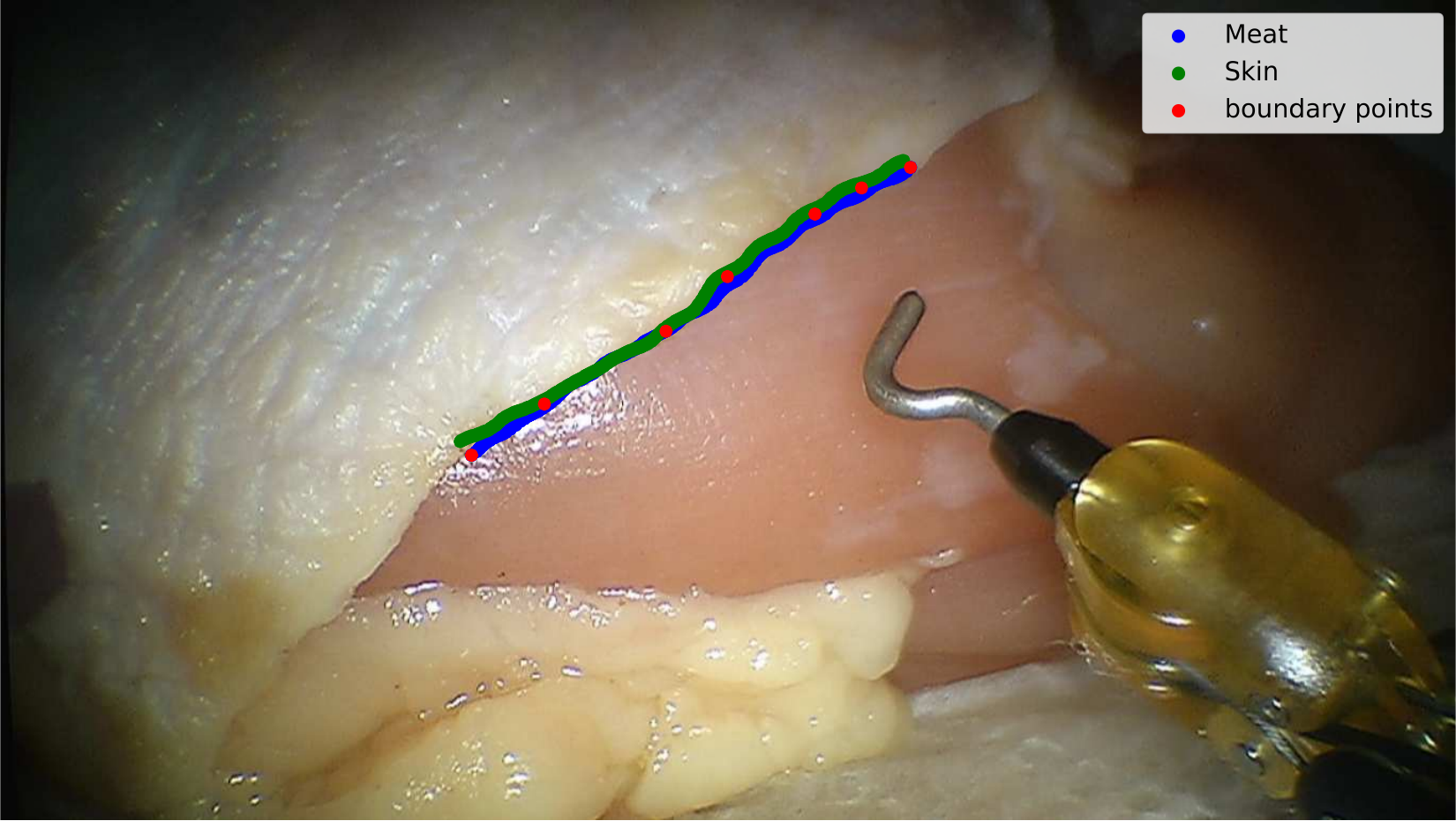}
        \caption{}
        \label{fig:chicken_bnd_pts2}
    \end{subfigure}
    \rulesep
    \hfill
    \begin{subfigure}[t]{0.3\textwidth}
        \centering
        \includegraphics[width=\linewidth]{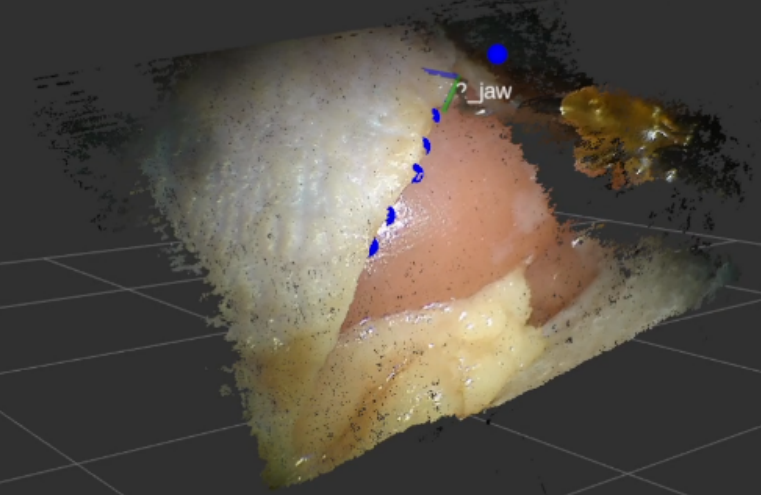}
        \caption{}
        \label{fig:chicken_pcl2}
    \end{subfigure}
    \rulesep
    \hfill
    \begin{subfigure}[t]{0.3\textwidth}
        \centering
        \includegraphics[height=0.2\textheight]{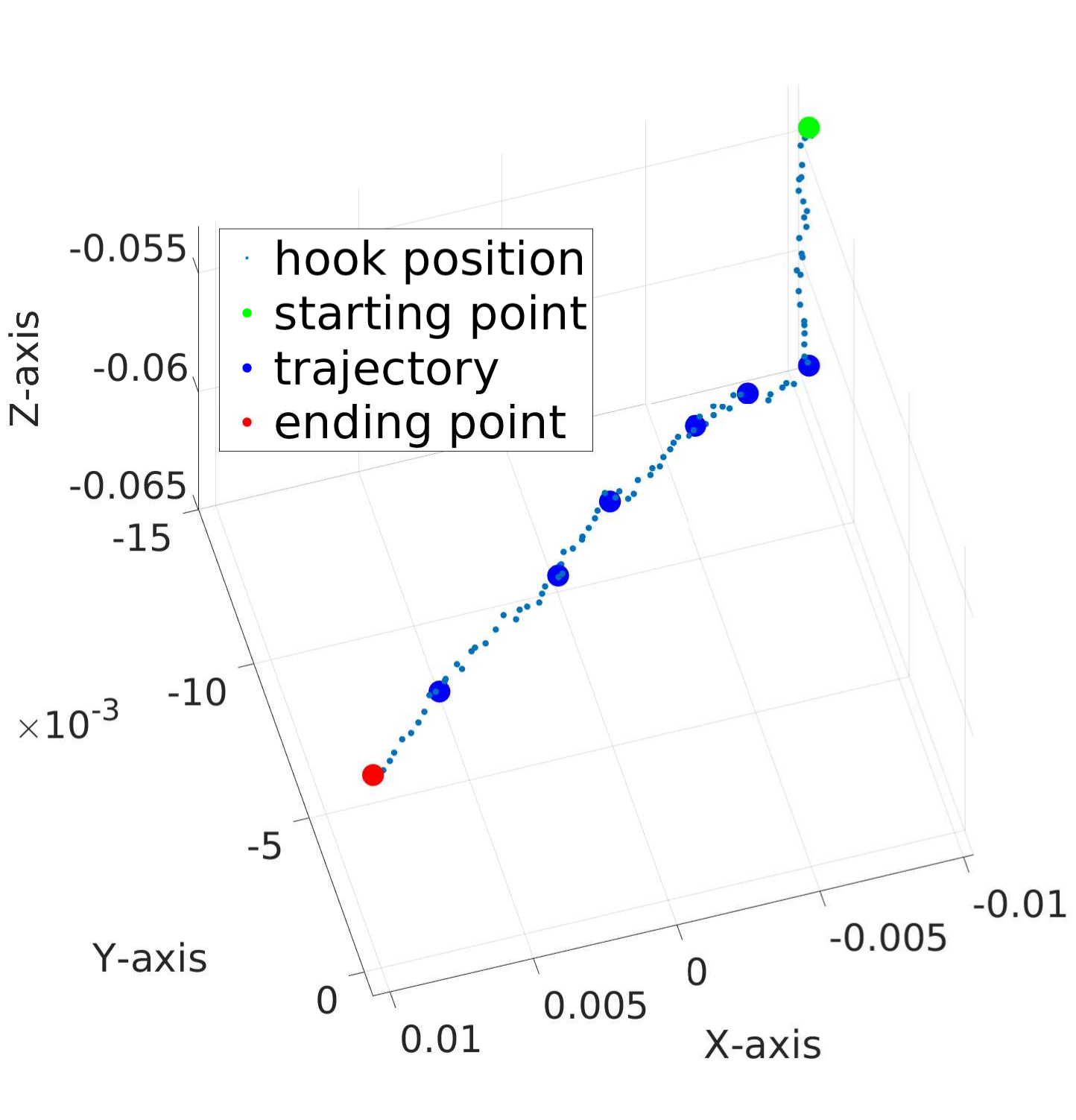}
        \caption{}
        \label{fig:chicken_traj2}
    \end{subfigure}

    \begin{subfigure}[t]{0.3\textwidth}
        \centering
        \includegraphics[width=\linewidth]{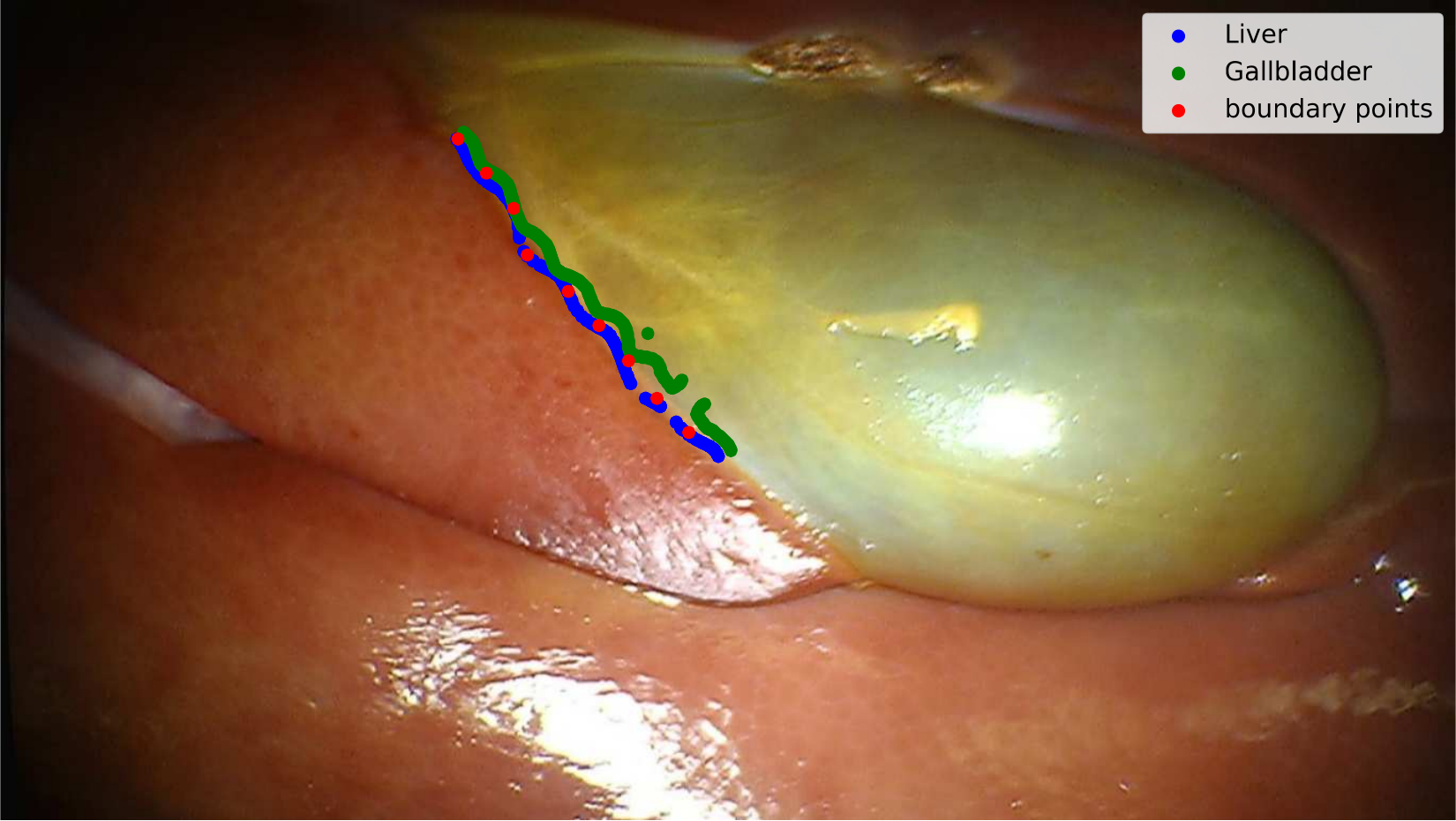}
        \caption{}
        \label{fig:liver_bnd_pts1}
    \end{subfigure}
    \rulesep
    \hfill
    \begin{subfigure}[t]{0.3\textwidth}
        \centering
        \includegraphics[width=\linewidth]{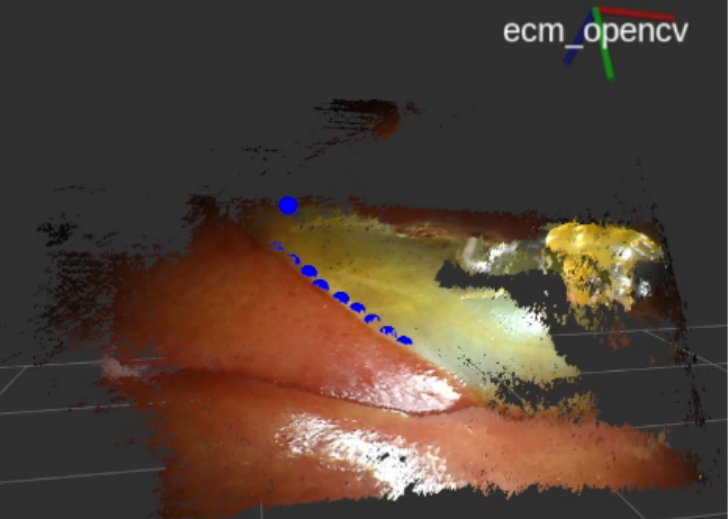}
        \caption{}
        \label{fig:liver_pcl1}
    \end{subfigure}
    \rulesep
    \hfill
    \begin{subfigure}[t]{0.3\textwidth}
        \centering
        \includegraphics[height=0.2\textheight]{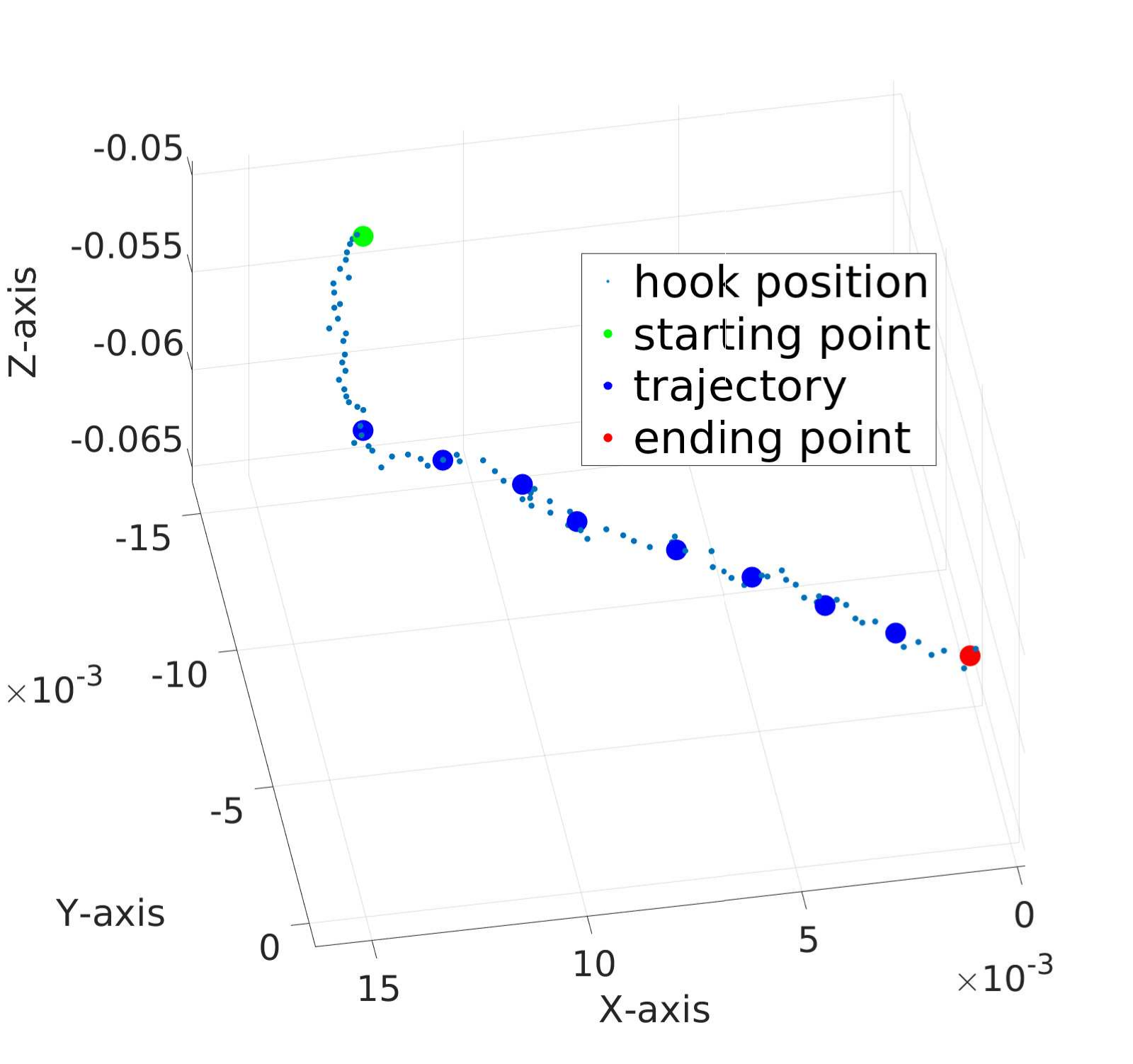}
        \caption{}
        \label{fig:liver_traj1}
    \end{subfigure}
    
    \begin{subfigure}[t]{0.3\textwidth}
        \centering
        \includegraphics[width=\linewidth]{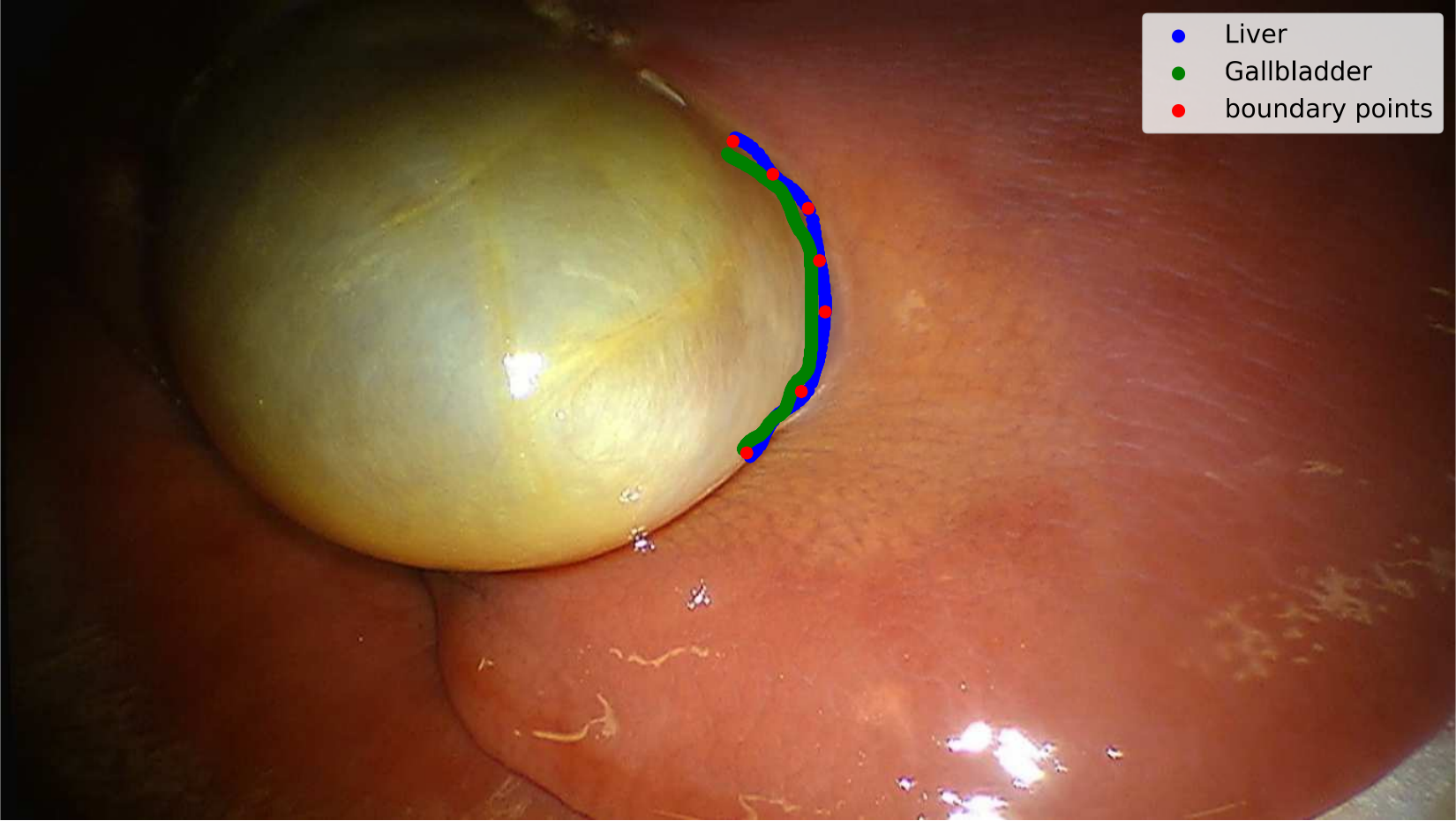}
        \caption{}
        \label{fig:liver_bnd_pts2}
    \end{subfigure}
    \rulesep
    \hfill
    \begin{subfigure}[t]{0.3\textwidth}
        \centering
        \includegraphics[width=\linewidth]{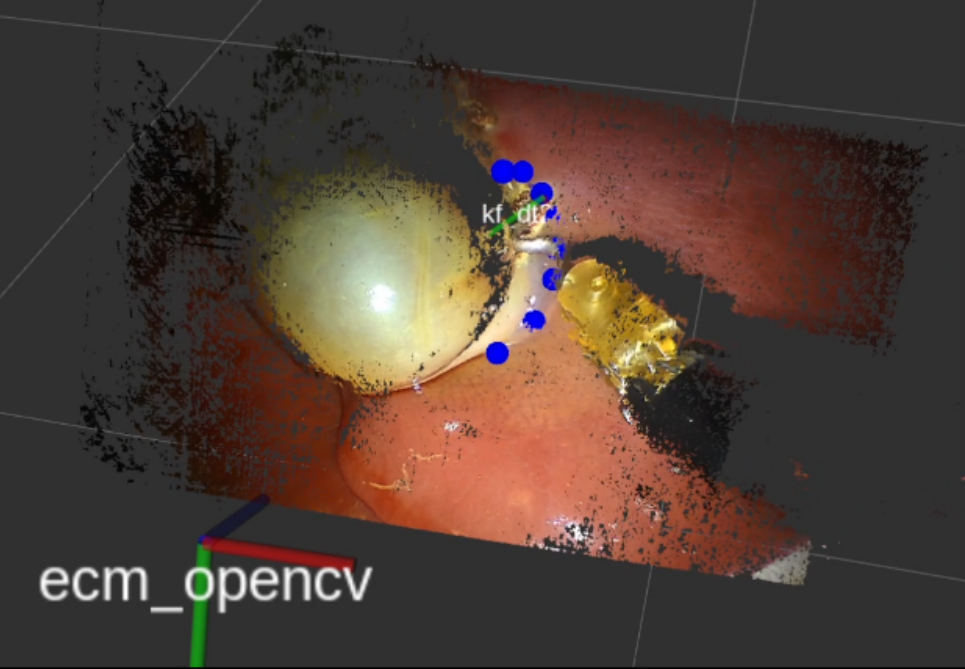}
        \caption{}
        \label{fig:liver_pcl2}
    \end{subfigure}
    \rulesep
    \hfill
    \begin{subfigure}[t]{0.3\textwidth}
        \centering
        \includegraphics[height=0.2\textheight]{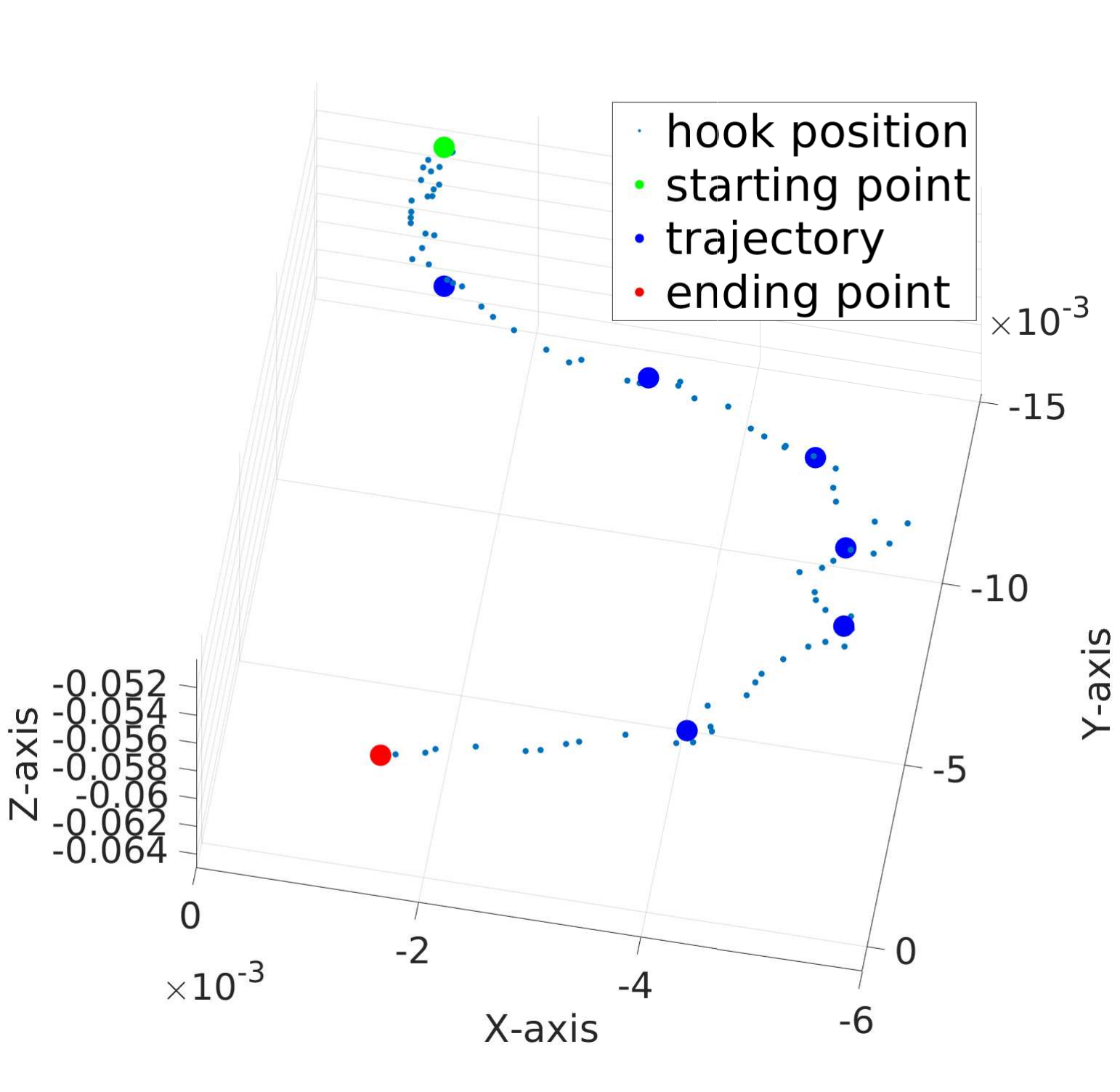}
        \caption{}
        \label{fig:liver_traj2}
    \end{subfigure}
    \caption{The left figures in each row ((a), (d), (g), and (j)) depict the final output of the image segmentation with the desired trajectory on the 2D endoscope image. The following points in the 3D space are shown in the middle figures ((b), (e), (h), and (k)). Finally, the right images ((c), (f), (i), and (l)) illustrate the instrument tip's movement trajectory during the procedure.}
    \label{fig:3dtrajs}
\end{figure*}

%% file: conclusion.tex
\section{Conclusion}



This paper describes a comprehensive framework for automated dissection using the da Vinci surgical robot, augmented with the da Vinci Research Kit (dVRK) and complemented by the Si model endoscope. This system integrates advanced robotics with AI-driven image-processing techniques, utilizing Detectron2, trained on our annotated dataset, for image segmentation and keypoint detection. Our ex-vivo evaluation with chicken and pig liver specimens demonstrates the system's capability in successfully dissecting target tissues, showing its potential in automating parts of surgical procedures like cholecystectomy. The results indicate remarkable performance in localizing surgical instruments and target tissues within endoscopic images.

Furthermore, the 3D scene reconstruction, achieved through the Semi-global Matching algorithm (SGM) on stereo endoscopic images, along with our fiducial marker-based calibration, has enabled real-time monitoring of the instrument's pose. This, in tandem with our feedback mechanism and control strategy, ensures the robot's precise trajectory following during the dissection process.

However, to fully dissect the gallbladder, it is imperative to address the following issues: augmenting the dataset for Detectron2, tracking the boundary of the tissues in real-time as they deflect, maintaining precise instrument tracking even when it is at the edges of the images, and introducing the second arm for grasping the gallbladder.

To resolve the issues, we collected a set of ex-vivo cholecystectomies performed by surgeons on our current da Vinci model~\cite{oh2023comprehensive}. Each trial was carried out with a different specimen, and diverse movements of the instruments were observed. The future model will be trained with this augmented dataset, expecting it to be rich enough to produce a real-time algorithm to robustly track the tissue boundary and plan instrument movement. This will allow the system to respond when the gallbladder is grasped and pulled or if the gallbladder deforms as a result of the energy delivery. Moreover, we will implement the endoscope movement as in~\cite{visualservo} to keep the instrument consistently centered. We also plan to initially combine manual control of the grasper with automated boundary tracking and energy delivery, as automating the grasping task is significantly more challenging than dissection. However, our long-term goal is to completely automate the dissection task with performance similar to that of a surgeon.


%% file: appendix.tex
\section{Appendix}

\subsection{Fiducial Marker Based Arm Calibration}
\label{sec:armcalib}

\input{tables/calib_joint_limits}

\input{tables/calibdata}

To explain the global optimization method used, we detail the robot joint movements during dataset recording, as outlined in Table~\ref{tab:jointlimit}. The procedure initiates with joint 1 and progresses through to the final joint, maintaining all other joints at a static position of $0$. Despite the robot arms not fully extending, the dataset, detailed in Table~\ref{tab:calibdata}, sufficiently captured the necessary information to estimate the joint twist parameters effectively.

For the Patient Side Manipulators (PSMs), the joint configuration string is RRPRRR, and for the Endoscopic Camera Manipulator (ECM), it is RRPR. In this configuration, R denotes a revolute joint, and P is a prismatic joint. The parameters for a revolute joint are defined as follows:
\begin{equation}
\xi^R = [\nu_x \quad \nu_y \quad \nu_z \quad \omega_x \quad \omega_y \quad \omega_z]^{T},
\end{equation}
where $\nu$ represents linear velocity, and $\omega$ rotational velocity. For prismatic joints, the parameters are given by:
\begin{equation}
\xi^P = [\nu_x \quad \nu_y \quad \nu_z \quad 0 \quad 0 \quad 0]^{T}.
\end{equation}
Following this, we adjusted the joint angle values as per Equation~(\ref{Eq:Angles}), leading to a total of 128 variables to be optimized — 48 per PSM and 32 for the ECM.

To measure the distance between two homogeneous transformation matrices indicative of rigid body motion, we use a weighted sum of translational (weight of $0.7$) and rotational distances (weight of $0.3$). The translational distance is calculated using the Euclidean distance, while the rotational distance is determined by:
\begin{equation}
\Delta\theta = 2\cos^{-1}(Re(z)),
\end{equation}
where $z = q_{1}\cdot\overline{q}_2$ (with $\overline{q}_2$ being the conjugate quaternion of $q_{2}$), and $q_{1}$ and $q_{2}$ represent the rotation quaternions of each transformation. The optimization objective is to minimize the mean squared error of these distances, subject to nonlinear constraints, ensuring that the norm of the rotational velocity is 1 and the last three values in the prismatic joint parameter vector are set to $0$.

\input{tables/calilbparams}

\subsection{Inverse Kinematics}

In our approach to numerical inverse kinematics, we implemented the Sequential Least Squares Programming (SLSQP) algorithm, a technique well-suited for minimizing scalar functions involving several variables within specified bounds and constraints~\cite{kraft1988software}. Our objective function, aimed at calculating the distance between two homogeneous transformation matrices, is detailed in Section~\ref{sec:armcalib}. We designed the system constraints to be adjustable, ensuring they remain within the joint limits of dVRK~\cite{dvrk}.




%% file: tables/calib_joint_limits.tex
\begin{table}[t]
\resizebox{\columnwidth}{!}{%
\begin{tabular}{cc|cccccc}
\hline \hline
\multicolumn{2}{c|}{\textbf{\begin{tabular}[c]{@{}c@{}}Joint\\ (metric)\end{tabular}}} & 1 (deg) & 2 (deg) & 3 (m) & 4 (deg) & 5 (deg) & 6 (deg) \\ \hline
\multicolumn{1}{c|}{\multirow{2}{*}{\textbf{ECM}}}  & Min & -70 & -45 & 0.03 & -85 & -   & -   \\
\multicolumn{1}{c|}{}                               & Max & 70  & 40  & 0.23 & 85  & -   & -   \\ \hline
\multicolumn{1}{c|}{\multirow{2}{*}{\textbf{PSMs}}} & Min & -90 & -45 & 0    & -60 & -30 & -80 \\
\multicolumn{1}{c|}{}                               & Max & 90  & 45  & 0.24 & 60  & 50  & 80  \\ \hline \hline
\end{tabular}%
}
\caption{The joint limit settings for each arm while recording the dataset. }
\label{tab:jointlimit}
\end{table}

%% file: tables/calibdata.tex
\begin{table}[t]
\centering
\begin{tabular}{c|c|c|c}
\hline \hline
\textbf{Arms}                & \textbf{PSM1} & \textbf{PSM2} & \textbf{ECM} \\ \hline
\textbf{Number of instances} & 1950          & 1950          & 1700         \\ \hline \hline
\end{tabular}
\caption{Number of recorded instances for calibration of each arm.}
\label{tab:calibdata}
\vspace{-5mm}
\end{table}

%% file: tables/calilbparams.tex
\begin{table*}[ht]
\centering
\begin{tabular}{c|cccccc|cccccc}
\hline \hline
          & \multicolumn{6}{c|}{\textbf{PSM1}}                        & \multicolumn{6}{c}{\textbf{PSM2}}                         \\ \hline
$\xi_1$   & 0.0002  & -0.0385 & -0.0967 & 0.9991  & 0.0265  & -0.0321 & -0.0028 & 0.0704  & 0.1640  & -0.9968 & -0.0775 & 0.0214  \\
$\xi_2$   & -0.1008 & 0.7068  & 0.0809  & -0.0348 & 0.1076  & -0.9936 & -0.0971 & -0.6964 & -0.0327 & -0.0254 & 0.0518  & -0.9983 \\
$\xi_3$   & 0.0262  & 0.9992  & 0.0286  & 0.0     & 0.0     & 0.0     & 0.0315  & 0.9995  & -0.0063 & 0.0     & 0.0     & 0.0     \\
$\xi_4$   & -0.0719 & 0.0745  & -0.7082 & 0.0517  & -0.9927 & -0.1093 & 0.0603  & 0.0391  & -0.6942 & -0.0458 & 0.9975  & 0.0529  \\
$\xi_5$   & 0.0662  & -0.7091 & -0.0787 & 0.0348  & -0.1076 & 0.9936  & -0.0574 & -0.6983 & -0.0352 & -0.0255 & 0.0518  & -0.9983 \\
$\xi_6$   & -0.0002 & -0.0425 & -0.0495 & 0.9991  & 0.0265  & -0.0321 & 0.0034  & -0.0719 & -0.1141 & 0.9968  & 0.0775  & -0.0214 \\
$\alpha$  & -1.0    & 1.0     & -1.0227 & 1.0     & -1.0    & -1.0    & -1.0    & -1.0    & -0.9998 & -0.9999 & -1.0    & 1.0     \\
$\beta$   & 0.0047  & 0.0289  & 0.0003  & 0.0     & 0.0288  & -0.0047 & 0.0031  & -0.0306 & 0.0005  & 0.0     & 0.0306  & 0.0031  \\ \hline
          & \multicolumn{6}{c|}{\textbf{ECM}}                         &         &         &         &         &         &         \\ \hline
$\zeta_1$ & 0.0019  & -0.0862 & -0.0753 & 0.9999  & 0.0159  & 0.0019  &         &         &         &         &         &         \\
$\zeta_2$ & -0.1008 & 0.7068  & 0.0809  & -0.0348 & 0.1076  & -0.9936 &         &         &         &         &         &         \\
$\zeta_3$ & -0.0051 & 0.9999  & -0.0130 & 0.0     & 0.0     & 0.0     &         &         &         &         &         &         \\
$\zeta_4$ & -0.0869 & 0.0276  & 0.5999  & 0.0011  & -0.9989 & 0.0478  &         &         &         &         &         &         \\
$\alpha$  &         & 1.0491  & -1.0134 & -1.0077 & 0.9815  &         &         &         &         &         &         &         \\
$\beta$   &         & 0.0017  & -0.0267 & 0.0302  & -0.0092 &         &         &         &         &         &         &         \\ \hline \hline
\end{tabular}
\caption{Optimized values for the unknown parameters.}
\label{tab:calibparams}
\vspace{-5mm}
\end{table*}